\documentclass{article}
\usepackage{iclr2027_conference}
\iclrfinalcopy   
\usepackage{times}
\usepackage{amsmath,amssymb,amsfonts}
\usepackage{graphicx}
\usepackage{booktabs}
\usepackage{multirow}
\usepackage{hyperref}
\usepackage{url}
\usepackage{microtype}
\usepackage{xcolor}
\usepackage{float}
\usepackage[all]{hypcap}
\title{The Signs Were Always There:\\
Training-Free Detection and Control\\
in Raw Transformer Dimensions}

\author{Varun Reddy Nalagatla \quad Madhavi Evana \quad Yiwen Zhang \\
Amazon Web Services}

\begin{document}

\maketitle

\begin{abstract}
Because of polysemanticity and superposition, it has been challenging to attribute token understanding and generation to specific neurons or dimensions. Yet the information in the hidden states is so rich that the signs of the residual stream can surface semantic content in early layers and control what the model generates. We call this sign-pattern-driven interpretation approach Bag of Dims (BoD).
Using BoD, we read each dimension of the residual stream individually. Since dimensions carry almost no shared information, a concept is simply the set of dimensions that agree in sign. This lets us discover tens to hundreds of thousands of concepts per model without training or labels. A concept is detectable as early as the first few layers, at AUC above 0.9 across the model families we tested, well before the answer it resolves to becomes legible. BoD can therefore screen harmful prompts at prefill, without a trained guard model.
The prediction capability of the signs is strong: setting all magnitudes to unity preserves 56--94\% top-5 next-token accuracy through the language model head. Flipping a concept's sign pattern during the forward pass suppresses its logits, and a separate training-free intervention steers a chosen concept into generated text.
The BoD method depends on the transformers' per-dimension sign pattern, present across the residual stream, attention, and MLP subspaces, rather than on any particular architecture, making it applicable to a wide range of models.

\end{abstract}

\section{Introduction}
\label{sec:introduction}

Mechanistic interpretability aims to reverse-engineer the computations performed
inside a trained neural network. A recent field-wide review emphasized the 
importance of research in this field for monitoring models for cognition related to dangerous behavior
\citep{sharkey2025openproblemsmechanisticinterpretability}. Individual neurons or hidden dimensions are
not directly interpretable, as superposition lets a network with $D$
dimensions represent many more than $D$ features by packing them into
overlapping, non-orthogonal directions \citep{elhage2022toy,elhage2023privileged}.
Reading a concept requires first finding the right
rotation of the basis. A line of research has looked into how to learn that rotation, including
fitting sparse autoencoders (SAEs), an overcomplete dictionary fit to millions of activations
\citep{cunningham2023sparse,bricken2023monosemanticity,templeton2024scaling},
probing a labeled concept one property at a time
\citep{alain2017understanding,belinkov2022probing}, using transcoders to approximate a
component \citep{dunefsky2024transcoders}, and steering activations
along a fitted direction
\citep{turner2023activation,panickssery2023steering,zou2023representation,li2023iti}. 
Each of these tools, dictionary, probe, transcoder, or steering vector, is
refit from scratch for every new property, layer, and model. However, SAEs trained on randomly initialized,
untrained transformers surface similarly clean-looking features, raising doubt
about how much of what they find reflects the trained model at all
\citep{makelov2025sparse}.

This paper presents a method that achieves concept detection without any training, and only leveraging the transformer hidden dimensions. The signs of the dimensions carry the content, while their magnitudes carry the strength. We define features as subsets of dimensions with consistent sign patterns among tokens representing the same concept. These features carry enough information to differentiate a concept from others. Overwriting every residual dimension
with just its sign and setting all magnitudes to unity still
preserves 56--94\% of top-5 next-token accuracy through the LM head, whereas permuting those same signs
across dimensions destroys it. We also found that concept detection (reading) and target-generation steering (writing) use different features. Moreover, the reading and writing features are nearly disjoint, with rank correlation 0.01--0.11. We call this sign-feature-based method \textbf{Bag of Dims} (BoD).

When we applied this finding to a live model, we looked not only at what the dimensions encode but at when a concept can first be detected across the model's depth, when the answer becomes readable, and where an intervention can land. We use this method in the LLM prefill and generation lifecycle to flag a risky prompt before generation, monitor an answer as it forms, and steer a target concept.

\section{Methodology}
\label{sec:method}

\subsection{The Bag-of-Dims framework}
\label{sec:discovery}

At any transformer block output layer, we read each of the $D$ residual dimensions on its own. Its sign
($\pm1$) can be used to interpret the semantic content; its magnitude is the
readout strength. A feature
is a subset of dimensions $\mathcal{T}\subseteq\{1,\dots,D\}$ with a consistent sign pattern $\boldsymbol{\pi}\in\{+1,-1\}^{|\mathcal{T}|}$ among tokens representing the same concept. Let $h_l$ represent the hidden state after $l$ transformer blocks and
$h_l[d]$ for its $d$-th dimension.

We build the cache by passing each token $t$ into the LLM on its own as $[\textsc{bos},t]$, and we read the hidden
state at the token's own position, after the \textsc{bos} prefix. We keep only its sign,
$\mathrm{sign}(h_l)$, one bit per hidden dimension, so each token becomes a $D$-bit sign pattern.
Presenting a token this way is context-free (Appendix~\ref{app:method}). For a concept with anchor tokens $\mathcal{A}_c$,
let $s_{l, t,d}=\mathrm{sign}(h_l[t,d])$ at the chosen layer $l$; we score each dimension by how well its
sign separates anchors from the full vocabulary $\bar{\mathcal{A}}_c$:

\begin{align}
p^+_d = \tfrac{1}{|\mathcal{A}_c|}\!\sum_{t\in\mathcal{A}_c}\!\mathbf{1}[s_{t,d}{=}{+}1],
\qquad
p^-_d = \tfrac{1}{|\bar{\mathcal{A}}_c|}\!\sum_{t\notin\mathcal{A}_c}\!\mathbf{1}[s_{t,d}{=}{+}1],
\label{eq:p_pos}
\end{align}

\begin{equation}
\mathrm{Score}_d = \max\!\Big(\tfrac{1+p^+_d-p^-_d}{2},\;\tfrac{1-p^+_d+p^-_d}{2}\Big).
\label{eq:per_dim_auc}
\end{equation}
The $\mathrm{Score}_d$ measures how well dimension $d$ differentiates the concept from the others.
Dimensions with $\mathrm{Score}_d\ge\kappa_{\text{reg}}{=}0.75$ are included in the detection feature, $(\mathcal{T}_c,\boldsymbol{\pi}_c)$, with polarity set by the sign of $p^+_d-p^-_d$.
A token $t$ is scored against feature $c$ by the fraction of the feature's dimensions
whose sign matches the expected sign of this concept,
\begin{equation}
a_c(t) = \frac{1}{|\mathcal{T}_c|}\sum_{d\in\mathcal{T}_c}\mathbf{1}\big[s_{t,d}=\pi_{c,d}\big],
\label{eq:agreement}
\end{equation}
and a feature's detection quality is the ROC AUC of $a_c$ separating the concept's tokens from the rest of
the vocabulary (reported as feature AUC throughout). A concept is reported only if it
clears a null of 100 random anchor sets of the same size.

\subsection{Unsupervised Discovery}
\label{sec:fracture}

In this section, we introduce a methodology to detect a concept at fine granularity and high accuracy. Given a seed token, we compute the distance between its sign vector $s_{l,t,d}$ and the sign vectors of a set of target tokens. Where that distance is small, the target token is considered a neighbor of the seed token, and the seed together with its nearest neighbors forms its \emph{sign-neighborhood}. For the seed ``France'', that
neighborhood mixes Paris and Rome, Spain and Germany, and the French words around
them.

The neighbors are semantically correlated with the seed token; however, they do not always
represent the same concept. We first collect the dimensions that do not have sign agreement, then run an unsupervised clustering algorithm on them. These clusters sort the neighbor tokens into clean groups that each represent a concept. For example, the neighbors of ``France'' are separated into its capital ($\{$Paris, Parigi, \dots$\}$) and its peer countries ($\{$Spain, Italy, \dots$\}$). We call this method fracturing.

Fracturing every seed token in turn yields a catalog where the seed tokens provide the labels.  We accept a cluster only if it passes a
parameter-free compression test---its shared sign pattern carries an excess
bit-deficit against a matched null of the same size---then merge concepts from
different seeds whose member sets overlap at Jaccard $\ge\tau_{\text{merge}}{=}0.90$,
which sets the catalog's granularity. The accepted $(\mathcal{T}_c,\boldsymbol{\pi}_c)$
reads a label-free concept and a curated one alike. Run in two sign spaces (the
embedding matrix, no forward pass; and the residual stream), this produces the eight
catalogs of \S\ref{sec:feature_discovery} and the dictionaries compared against
trained SAEs in Appendix~\ref{app:unsupervised}. The full protocol and the subspace
definitions are in Appendix~\ref{app:method}.

\subsection{The write catalog}
\label{sec:method_write}

A dimension's read role differs from its
write role (how strongly its sign drives a concept's output). The write
target for concept $c$ with seed set $\mathcal{S}_c$ is the per-dimension
consensus of the unembedding over the seeds,
\begin{equation}
\boldsymbol{\tau}_c = \mathrm{sign}\!\Big(\textstyle\sum_{t\in\mathcal{S}_c} W_{\mathrm{unembed}}[t]\Big),
\label{eq:write_target}
\end{equation}
where $W_{\mathrm{unembed}}$ is the model's unembedding matrix (the LM head), and
$W_{\mathrm{unembed}}[t]$ its row for token $t$, the output direction that token's
logit reads from. The target $\boldsymbol{\tau}_c$ is a property of the frozen
weights, needing no forward or backward pass; the seeds
are the concept's own members from the same label-free fracture
(\S\ref{sec:fracture}), here applied to a single concept word. Read and write are
defined on the same basis, but the dimensions they use turn out to be nearly
disjoint (\S\ref{sec:write_catalog}); how $\boldsymbol{\tau}_c$ is delivered into a
running forward pass to steer generation is taken up with the control results
(\S\ref{sec:write_catalog}).

\section{The sign-based concept detection}
\label{sec:experiments}

All experiments in this section use four publicly available language models:
Qwen 3.5-4B, Gemma 3-4B-pt, Mistral 7B v0.3, and Qwen3-32B. The single-token
hidden states are computed once per model and used in all the analyses; the vision and
audio transformers are introduced with the cross-modal results
(\S\ref{sec:cross_modality}).

\subsection{Sign Encodes Content, Magnitude Encodes Confidence}
\label{sec:sign_content}

Setting all magnitudes of the final hidden state to $1$ and reading only the sign through the LM head 
preserves 56--94\% of top-5 next-token accuracy across the four models. 
In contrast, the top-5 next-token accuracy dropped to $0\%$ for a random sign permutation (Appendix Table~\ref{tab:sign_prediction}). Restricting this analysis to only the 800 highest-magnitude
dimensions matches or beats using all dimensions. 
Magnitude marks which dimensions carry signal while sign encodes the content.

\subsection{Concepts Are Detected at Every Depth}
\label{sec:layer_invariance}

A feature is registered on the type cache, and the type cache can be built at any
layer $l$ using $s_{l,t,:}$, so detection is not tied to a particular depth. Sweeping all layers of the
four models on the 175 concepts, we find the average per-dimension score stays within 0.848--0.915
(end-to-end variation $<0.056$), with 98--100\% of concepts discoverable at every
depth, so no layer search is required, unlike a per-layer trained dictionary. Full per-layer
curves and the label-free discovery yield, which do vary with depth, are
in Appendix~\ref{app:layers}.

\subsection{Zero-Training Feature Discovery}
\label{sec:feature_discovery}

We apply the discovery method (\S\ref{sec:method}) to 175 concepts spanning animals, emotions, numbers, code constructs, grammar, professions, etc. Each category has 50 hand-curated anchor tokens. The negative set is the full vocabulary. Table \ref{tab:cross_model} has outputs from the experimentation. 

The concepts were detected with high AUC (0.995--1.000). All 175 concepts (171 on Mistral) exceed the null calibration threshold on every model despite different training data, vocabulary sizes, and dimension counts. Per-dim score measures how well individual dims can detect this category, and AUC measures concept detection accuracy.

\begin{table}[H]
\centering
\caption{Cross-model feature discovery (175 categories, 50 anchors each, full-vocabulary negatives). Per-dim AUC: best single dimension's separability. AUC: composite sign-agreement classifier using all dims above the 0.75 registration threshold (0.70 fallback when no dim clears 0.75).}
\label{tab:cross_model}
\resizebox{\textwidth}{!}{%
\begin{tabular}{llccccc}
\toprule
Model & Sign space & Discoverable ($\geq 0.75$) & Per-dim score & Per-dim null & Feature AUC & Exceed null \\
\midrule
Gemma 3-4B   & $h_{26}$   & 175/175 & 0.877 & 0.631 & 0.9989 & 175/175 \\
             & embedding & 175/175 & --- & --- & 0.9996 & 175/175 \\
\midrule
Qwen 3.5-4B  & $h_{24}$   & 175/175 & 0.874 & 0.646 & 0.9997 & 175/175 \\
             & embedding & 175/175 & --- & --- & 0.9996 & 175/175 \\
\midrule
Qwen3-32B    & $h_{48}$   & 175/175 & 0.871 & 0.648 & 0.9992 & 175/175 \\
             & embedding & 170/175 & --- & --- & 0.9953 & 175/175 \\
\midrule
Mistral 7B   & $h_{24}$   & 168/171 & 0.854 & 0.651 & 0.9983 & 171/171 \\
             & embedding & 163/171 & --- & --- & 0.9984 & 171/171 \\
\bottomrule
\multicolumn{7}{l}{\footnotesize Per-dim score is the best single dimension per concept, so its chance level is an extreme-value} \\
\multicolumn{7}{l}{\footnotesize statistic over $D$, not 0.5; the null column is the 95th percentile over 100 random 50-token anchor} \\
\multicolumn{7}{l}{\footnotesize sets. Mistral's 32K vocabulary resolves 171 of the 175 concepts to single tokens.} \\
\end{tabular}}
\end{table}

For each of the four models, we run the unsupervised procedure of \S\ref{sec:fracture} in two independent sign spaces, the embedding matrix (no forward pass) and the residual stream (one forward pass per vocabulary token). For each concept, we take a set of tokens as anchor tokens, and then measure with AUC whether its own members can be detected against random negative vocabulary tokens. Redundancy is the mean Jaccard similarity between concepts' token sets, so $0.000$ means concepts are not duplicates of one another. We also report the median number of dimensions in a feature, $|\mathcal{T}_c|$, and the median number of tokens in a member set.

\begin{table}[H]
\centering
\caption{Unsupervised discovery}
\label{tab:unsupervised}
\resizebox{\textwidth}{!}{%
\begin{tabular}{llccccc}
\toprule
Model & Sign space & Discovered & Median dims & Median members & AUC & Redundancy \\
\midrule
Gemma 3-4B    & embedding    & 480{,}798 & 41  & 18 & 0.999 & 0.000 \\
Gemma 3-4B    & $h_{26}$     & 422{,}645 & 44  & 24 & 0.998 & 0.000 \\
Qwen 3.5-4B   & embedding    & 400{,}164 & 32  & 19 & 0.998 & 0.000 \\
Qwen 3.5-4B   & $h_{24}$     & 682{,}323 & 137 & 18 & 1.000 & 0.000 \\
Mistral 7B    & embedding    &  37{,}327 & 18  & 17 & 0.990 & 0.000 \\
Mistral 7B    & $h_{24}$     &  81{,}006 & 92  & 17 & 0.999 & 0.000 \\
Qwen3-32B     & embedding    & 140{,}500 & 18  & 21 & 0.986 & 0.000 \\
Qwen3-32B     & $h_{48}$     & 286{,}123 & 181 & 26 & 0.998 & 0.000 \\
\bottomrule
\end{tabular}}
\end{table}

\subsection{Features Generalize to Context and Act Causally There}
\label{sec:contextual}

Concept-defining features, discovered from isolated tokens, detect the same
concept in running text (mean contextual AUC 0.82--0.96 over 25{,}600 tokens);
about 58\% of dimensions keep their concept sign, and the flipping 42\% are
low-magnitude dims below the feature threshold. A sharper test uses polysemous
words whose senses straddle a concept boundary: the concept feature scores
the word higher in its context than in its other-sense context in
69--83\% of cross-concept cases (40--70 per model) on all four models (Appendix
Table~\ref{tab:polysemy}); contextual meaning is read from the same per-dim
feature, with no training.

The signs are causally operative, not merely readable. Flipping a feature's
registered signs \emph{away} from expected during a live forward pass lowers the
concept's target logits (mean change $-5$ to $-22$ across models), while forcing
them \emph{toward} expected, flipping a disjoint coalition, or flipping random
dimensions all do nothing (Appendix Table~\ref{tab:causal_flip}). Three controls
isolate the cause: sign not magnitude (toward $\approx 0$), the coalition not the
single dimension (the effect switches on only past $\sim$200--500 flipped dims),
and concept-specificity (a disjoint coalition is inert). The edit can only
suppress a concept; inducing it requires the write catalog.

\subsection{Features Survive Attention Projections}
\label{sec:attention}

Running the same 175-concept discovery on the K and V projection dimensions
(captured through the full compute path including RoPE) detects every concept
above null on both K and V across all four architectures, and likewise on the MLP
gate activation (Appendix Table~\ref{tab:kv}; feature-level comparison with the residual
stream in Appendix~\ref{app:ffn_signs}). The $W_k$ and $W_v$ projections
preserve the axis-aligned structure; the same holds for vision and audio
(\S\ref{sec:cross_modality}).

\section{The Dimensions that Control Writing}
\label{sec:write_catalog}

The suppression above is asymmetric: flipping read signs \emph{away} suppresses,
but \emph{toward} does not induce, because the read dimensions already carry the
expected sign. Induction lives elsewhere. The write target $\boldsymbol{\tau}_c$
(Eq.~\ref{eq:write_target}), a per-dimension consensus read off the frozen
unembedding with no forward or backward pass, is nearly disjoint from the read
catalog (rank correlation 0.01--0.11, falling to 0.008 on the 32B) and
concept-specific by construction (cross-concept Jaccard 0.02--0.07; Appendix
Table~\ref{tab:write_catalog}). Detection and production are carried by different
dimensions of the same basis.

Delivery is simple. We do not overwrite the residual read address; instead we add
the write target to the input of the attention output projection $W_O$
(\texttt{o\_proj}) across a mid-stack band, then watch for the concept to become
present and release the write once it does. This induces a concept in fluent text
$60$--$77\%$ of the time across the four models, training-free and with a
logit-clean decode (no repetition penalty or $n$-gram blocking; scored by an LLM
judge for both concept presence and fluency against the unsteered baseline). The target, not the prompt, drives the output: on Gemma
the single prefix ``Yesterday I went to the \dots'' continues to a supermarket
\textbf{(food)}, to ``the hospital to see my mom. She was in a lot of pain''
\textbf{(emotion)}, or to ``the doctor for a check up on my knee'' \textbf{(body)}, same decode,
differing only in $\boldsymbol{\tau}_c$. Per-model settings (shared across all
twelve concepts), the coalition and release rule, and the LLM-judge protocol are
in Appendix~\ref{app:method} (Table~\ref{tab:control}); per-concept generations and
the failures at the presence--coherence frontier are in
Appendix~\ref{app:write_catalog}. The closest precedent, Inference-Time Intervention
\citep{li2023iti}, also writes in the $W_O$-input space but derives its direction
from trained probes rather than the weight-sign consensus.

\section{Cross-Modality Universality}
\label{sec:cross_modality}

To test whether per-dim sign structure is specific to language, we apply the
identical procedure (\S\ref{sec:method}) to three non-language transformers:
\textbf{DINOv2-Base} \citep{oquab2024dinov2} (self-supervised vision, no
labels), \textbf{ViT-Base} \citep{dosovitskiy2021image} (supervised vision,
ImageNet-1K), and \textbf{AST} \citep{gong2021ast} (supervised audio, evaluated
on ESC-50 \citep{piczak2015esc}). DINOv2 and ViT-Base share one architecture and
evaluation set and differ only in training objective, isolating the effect of
supervision.

The same structure appears in every modality (summary table in Appendix~\ref{app:generality}).
It is present \emph{without} supervision (DINOv2 detects 9/12 ImageNet
superclasses from signs alone, feature AUC 0.831) and sharpened \emph{by} it
(ViT-Base 11/12, 0.977); audio is strongest (AST 50/50 concepts above 0.70,
47/50 above 0.80). Reading each dimension alone again loses nothing: a
full-capacity MLP adds at most $+0.007$ AUC over axis-aligned logistic
regression (exactly $0$ on the supervised models), pairwise MI stays at
0.001--0.002 bits, and pure sign Hamming retrieval reaches 93--97\%. The K/V
projections preserve the structure on all three models, as in language
(\S\ref{sec:attention}).

Both catalogs port. Reading a class's write target off the frozen classifier
head, $\boldsymbol{\tau}_c = \mathrm{sign}(W_{\mathrm{cls}}[c])$, is disjoint
from the read catalog (rank correlation 0.01--0.02, matching language) and,
injected through the same attention output pathway (\S\ref{sec:write_catalog}), drives
the classifier decision to the target class at 100\% capture against 0\% for a
sign-shuffled target of identical size. Detection and control are thus
modality-general, not language-specific. Full per-concept tables, the K/V
analyses, the depth-gated disjointness curve, and the directed-control
controls (specificity, amplifier, sub-band) are in
Appendix~\ref{app:generality}.


\section{The Feature Lifecycle}
\label{sec:lifecycle}

To read a hidden state as the model computes it, at any depth, we introduce the
\textbf{Sign Lens}, a training-free instrument and the only reader used in this
paper.
\label{sec:signlens}
The live state is the query and that layer's \emph{type cache} supplies the
references, the cached sign pattern of every vocabulary token read at the same
depth (\S\ref{sec:discovery}): the lens names what the state is carrying by
\emph{sign agreement}, the fraction of dimensions whose sign matches a cached
pattern, with magnitude at most a vote weight. Because query and references are
both $\mathrm{sign}(h_l)$, they share one frame and nothing needs per-layer
fitting, unlike the logit and tuned lenses, which read every intermediate state in
the model's final output frame and fit a per-layer affine probe to correct the
resulting bias \citep{belrose2023eliciting}. One instrument then supports three
reads, differing only in candidate scope and whether the neighborhood is clustered:
\textbf{decode} ranks a task-supplied answer set and reports the winner's margin,
used where the answer space is known (the correctness monitor,
\S\ref{sec:monitor}); \textbf{neighbors} ranks the whole vocabulary and reads the
composition of the top-$N$ neighborhood, used for concept and policy detection
with no candidate set (the prefill guardrail, \S\ref{sec:guardrail}); and
\textbf{fracture} splits that neighborhood into concept groups, used for
label-free discovery and for reading the lifecycle itself
(\S\ref{sec:detect-before-read}). Exact-token reading is candidate-restricted
because an open-vocabulary neighborhood does not reliably contain one specific
answer token, whereas concept detection needs no candidate set and stays
label-free.

\subsection{Detectability precedes readability}
\label{sec:detect-before-read}

Tracing the same sign addresses through depth reveals a consistent ordering: a
feature's concept is present in the signs early, while the specific answer it
resolves to becomes readable only late. Two complementary measurements bracket
the effect, and the Sign Lens makes it directly legible. Both hold on all eight
models: the four language models of \S\ref{sec:experiments} plus Mamba-2.8B (SSM),
OLMoE-1B-7B (mixture of experts), Pythia-1.4B, and RWKV-4-3B (linear-attention RNN).

\paragraph{The concept is written early (causal).}
We corrupt a subject token and restore only the clean \emph{sign} pattern at a
single early layer, keeping the corrupted magnitudes. Restoring signs recovers a
median 94--100\% of the destroyed answer probability across the eight models over the
first 15\% of depth, while restoring \emph{random} signs pushes the answer below its
corrupted level: the answer content is carried by the sign pattern and is present in
the earliest layers (curves in Appendix~\ref{app:circuit}).

\paragraph{The answer is read late (correlational).}
Decoding the same signs by counting sign agreements against each candidate's
concept address reaches 68--80\% top-1 accuracy, but only in the late layers
(peak at 74--89\% depth); magnitude weighting sharpens the onset without changing
the frame (Section~\ref{sec:sign_content}). The gap between the early write and
the late read is the lifecycle, plotted on a shared depth axis in
Appendix~\ref{app:circuit} (Figure~\ref{fig:early_late}).

\paragraph{Reading concepts off the signs.}
The ordering is legible with no trained reader at all. Following the prompt
\emph{the capital of France is} down the stack with the Sign Lens, the
last-position neighborhood reads \texttt{capital, city, currency} by 12\% of
depth, holds candidate capitals \texttt{London, Madrid, Paris} by 38\%, and only
near 75\% does \textbf{\texttt{Paris}} itself dominate. The concept arrives
early and the answer late, the same ordering the two measurements above give,
read straight off the signs (a second prompt and the full neighbor sets are in
Appendix~\ref{app:evidence}).

\section{Lifecycle-Aware Applications}
\label{sec:applications}

Each application is one read of the Sign Lens (\S\ref{sec:signlens}) at the layer
the lifecycle prescribes: the guardrail reads the neighborhood early, the monitor
decodes a candidate set late, and steering writes mid-stack. The guardrail and
monitor are evaluated on eight models spanning architectures: the four language
models of \S\ref{sec:experiments} plus Mamba-2.8B (SSM), OLMoE-1B-7B (mixture of
experts), Pythia-1.4B, and RWKV-4-3B (linear-attention RNN). Steering is reported
on the four language models.

\subsection{Prefill guardrail}
\label{sec:guardrail}

The guardrail classifies a prompt at the last prefill token, before generation,
inheriting the early detectability of \S\ref{sec:detect-before-read}.

\paragraph{Training-free gating.}
We name the harm concept with a few anchor words (weapon and violence terms)
against a benign background, read their sign address from the type cache, and
score a prompt by sign agreement at the last prefill token, with no labeled
corpus and no fitting. On all eight models this gate separates harmful from
benign prompts at AUC 0.93--0.99 within the first 12\% of depth, and on the harder
hard-negative contrast (benign prompts borrowing charged wording, such as ``kill a
runaway process'') still reaches 0.86--0.95 by 19\% depth on seven of the eight
(RWKV-4-3B reaches 0.79; Table~\ref{tab:harm_depth}): the intent, not merely the
charged token, is legible at prefill.

\paragraph{Calibration on an approved benchmark.}
To confirm the signal lives in the signs, we compare two \emph{supervised}
readouts on RealToxicityPrompts \citep{gehman2020realtoxicityprompts}, fit on the
same labeled split: a closed-form count over the sign bits (a class-conditional
Bernoulli vote, no gradient and no learned direction) and a trained linear probe
on the raw activations. The sign count uses the labels but adds no learned
parameters, and across the eight models it peaks at 0.90--0.93 against a
probe ceiling of 0.91--0.95, already reaching 0.80--0.89 by 12--17\% of depth (per-model values in
Appendix~\ref{app:evtables}). Tracking a fitted geometry to within a few points
with no learned direction locates the toxicity signal in the sign bits themselves;
and the count is auditable, its highest-weighted Mistral dimensions decoding to a
coherent profanity neighborhood.

\subsection{Correctness monitor}
\label{sec:monitor}

The read-late readout supplies a confidence signal at no extra cost: at a late
layer we score each candidate answer by sign agreement and take the \emph{margin}
between the top candidate and the runner-up. This margin is calibrated: across the
eight models it separates correct from incorrect readouts at AUC 0.73--0.98, and
keeping the most-confident half of prompts raises accuracy from a 0.67--0.80 base
to 0.85--1.00 (Appendix Table~\ref{tab:monitor}). The monitor reads the same
late-layer signs the answer is written into, so it can abstain as a low-confidence
answer forms, with no probe and no training.

\paragraph{Scope.}
On this battery the model answers correctly on 95--100\% of prompts, so the
monitored quantity is the correctness of the sign readout over the closed
candidate set (selective prediction), not hallucination; a sign-native
hallucination monitor is left to future work.

\subsection{Concept steering}

The third application writes rather than reads: at a mid-stack band it induces a
named concept by driving the write target $\boldsymbol{\tau}_c$ into the
attention-output projection, the intervention site the lifecycle prescribes. The
mechanism and results (fluent, training-free induction $60$--$77\%$ of the time
across four models) are given with the write catalog in \S\ref{sec:write_catalog};
delivery is set per model because a single global strength over-drives some
architectures and under-drives others.

\section{Related Work}
\label{sec:related}

\S\ref{sec:introduction} placed this work against the rotate-then-read pipeline;
here we record what separates it from each line specifically.

\textbf{Learned dictionaries.} Sparse autoencoders
\citep{bricken2023monosemanticity,templeton2024scaling,cunningham2023sparse} are
motivated by superposition \citep{elhage2022toy}, features at non-orthogonal
directions recoverable only through a learned rotation; the combinatorial capacity
of sign patterns removes that pressure, and we match a trained SAE from the
standard basis (Appendices~\ref{app:typelevel}--\ref{app:tpp}). Dictionaries
trained on vision \citep{joseph2025steering} and audio \citep{paek2025learning}
have the same per-model cost our cross-modal results avoid. That SAEs yield clean
features even on untrained models \citep{makelov2025sparse} does not carry over
here: at random init per-dim concept score collapses to 0.60--0.68 with 0--11/175
detectable, so the structure requires training (Appendix~\ref{app:trained_random}).

\textbf{Reading intermediate states.} The logit and tuned lenses
\citep{nostalgebraist2020logit,belrose2023eliciting} read every state in the
model's \emph{output} frame, the latter fitting a per-layer affine probe to correct
the resulting bias; the Sign Lens decodes each layer in its \emph{own} frame, so
nothing needs correcting and nothing is trained (\S\ref{sec:signlens}).
Probes \citep{alain2017understanding,belinkov2022probing} are fit per property, and
ours converges to axis-aligned voting rather than a rotation. That the residual
stream is empirically axis-aligned despite being theoretically rotation-invariant
\citep{elhage2023privileged} is the precondition a sign decoder exploits, and the
axis-versus-distributed question has a long vision literature
\citep{bau2017network,olah2017feature,fong2018net2vec,kim2018tcav}. All of it reads
a unit's \emph{magnitude}, and trains a decoder to do so; we ask instead whether a
coordinate's \emph{sign} suffices, untrained.

\textbf{Training-free steering.} ActAdd \citep{turner2023activation} and
contrastive activation addition \citep{panickssery2023steering} build a direction
from contrastive prompts, Inference-Time Intervention \citep{li2023iti} from
trained probes in the attention-output space, and RepE
\citep{zou2023representation} finds concept directions for monitoring and control.
Each yields a bespoke artifact unrelated to any detector. Ours is read from the
frozen unembedding and is the \emph{same} sign object the read side uses, so
detection and control become two uses of one weight-derived quantity.

\section{Discussion}
\label{sec:discussion}

\textbf{Combinatorial, not geometric, capacity.} With $D$ dimensions the sign
patterns span $3^D-1$ possible features, so the concepts we discover need not
compete for orthogonal directions: scoring each on its own registered dimensions
lets them share dimensions without interfering (member overlap $\le0.001$). This
does not refute superposition, which per-dim sign matching also decodes in
Elhage et al.'s toy model; reading one dimension at a time is a \emph{functional}
independence claim, not a statistical one
(Appendices~\ref{app:superposition},~\ref{app:discussion}).

\textbf{Limitations.} We test 1.4--32B language and 86M vision/audio models (70B+ open); suppression is
validated at the logit level, not as fluent generation; and steering uses one
shared per-model setting, so resistant concepts are reported rather than tuned.
The write target needs an output head, excluding DINOv2 from control. Full
discussion is in Appendix~\ref{app:discussion}.

\section{Conclusion}
\label{sec:conclusion}

Individual dimensions act as binary registers read one at a time: their signs encode content, their magnitudes strength. Sign alone carries 56--94\% of top-5 accuracy, label-free discovery yields 37{,}000--682{,}000 concepts per catalog, and detection and control occupy nearly disjoint, training-free dimensions across language, vision, and audio, shifting the open problem from finding the right rotation to cataloging what each dimension encodes.

\bibliography{references}
\bibliographystyle{iclr2027_conference}

\appendix

\section{The Lattice, and Why We Read Dimensions Individually}
\label{app:lattice}

\begin{figure}[h]
\centering
\includegraphics[width=\textwidth,height=0.21\textheight,keepaspectratio]{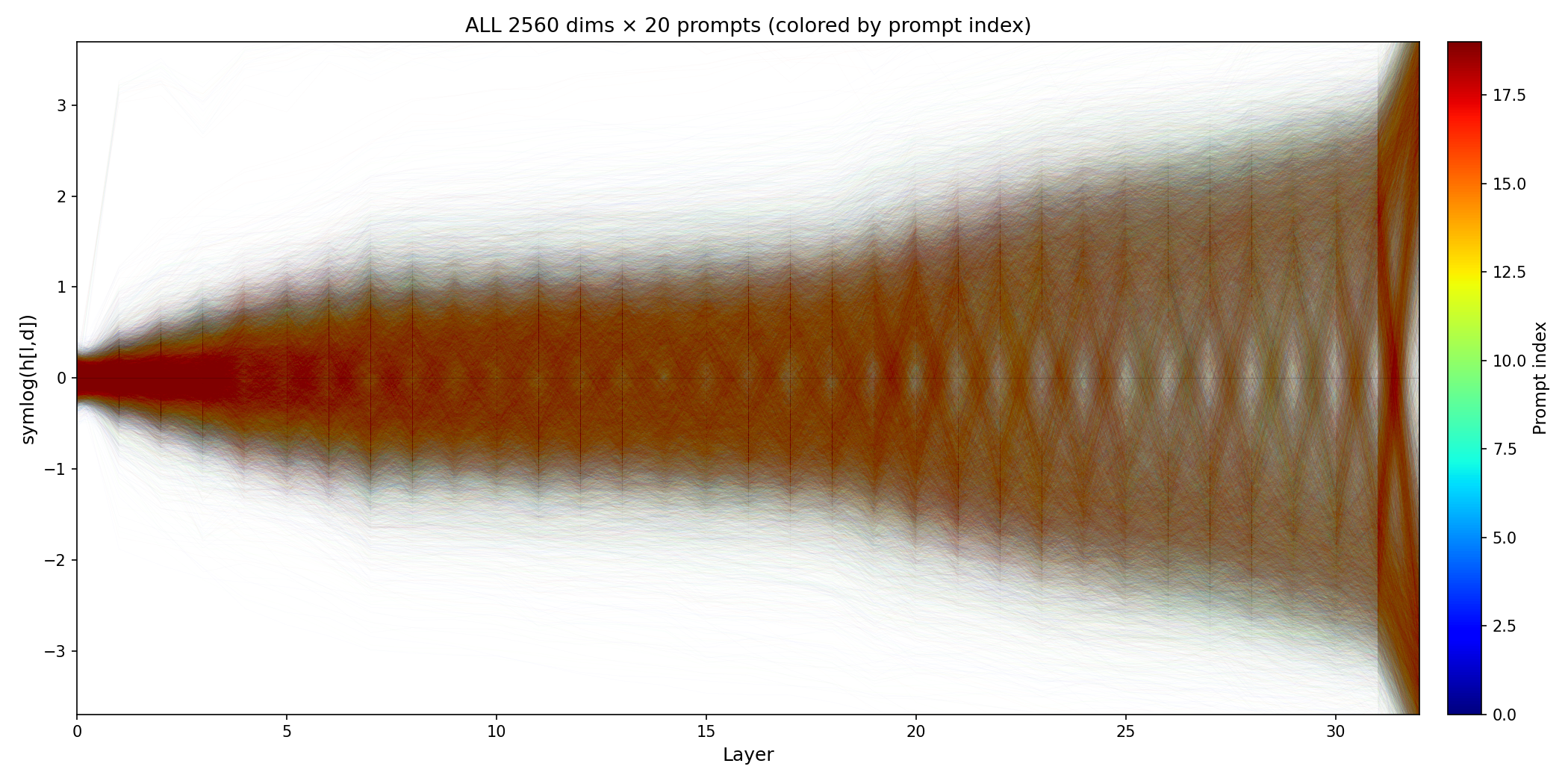}
\caption{Twenty different prompts overlaid (Qwen 3.5-4B). All 2560 dimensions of the
residual stream state $h_l[d]$ (the layer output after $l$ blocks) plotted across
layers. The same expanding structure appears regardless of input content.}
\label{fig:all_prompts}
\end{figure}

This appendix records the observation that motivated the per-dimension reading.
Plotting every dimension of a hidden state across layers, with twenty prompts
overlaid, shows a repeating pattern: a banded diamond lattice that expands with
depth and looks the same whatever the input (Figure~\ref{fig:all_prompts}).

\begin{figure}[h]
\centering
\includegraphics[width=\textwidth,height=0.42\textheight,keepaspectratio]{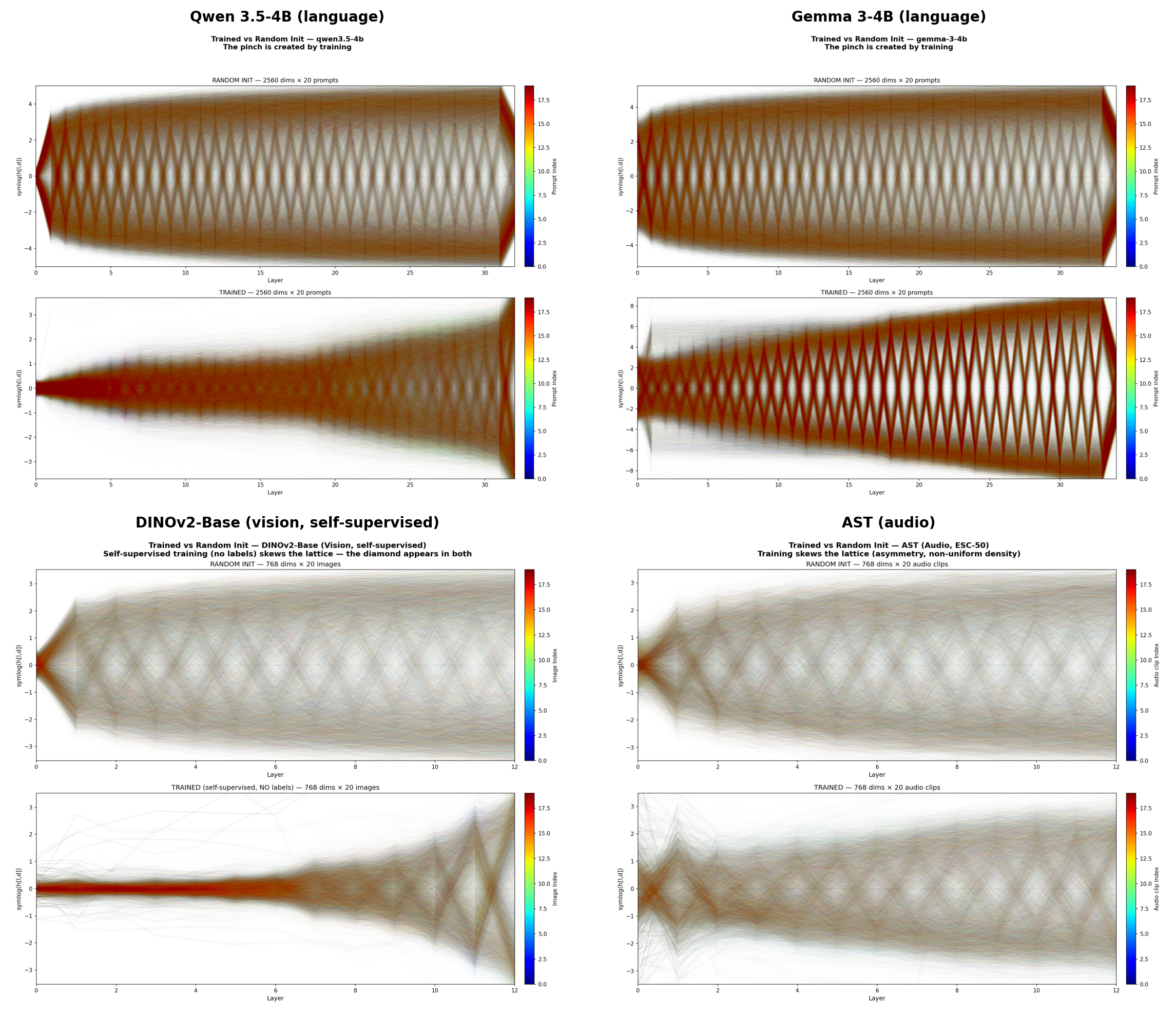}
\caption{Trained (bottom of each panel) vs.\ random init (top) across language
(Qwen 3.5-4B, Gemma 3-4B), vision (DINOv2-Base, self-supervised), and audio (AST).
The expanding envelope is a population artifact; the internal structure (pinch,
banding, non-uniform density) is created by training.}
\label{fig:trained_vs_random}
\end{figure}

The first thing to rule out was that training produced it. Running the same inputs
through randomly initialized models---in language, vision, and audio---the lattice
is still there (Figure~\ref{fig:trained_vs_random}). So the lattice is not something
training creates; what training creates is the structure \emph{inside} it: random
weights give a symmetric lattice of uniform density, while trained weights skew it
into a characteristic pinch, with asymmetric banding and non-uniform density, in
every modality (the feature-level consequence is quantified in the random-init
control, Appendix~\ref{app:trained_random}).

\begin{figure}[h]
\centering
\includegraphics[width=0.76\textwidth]{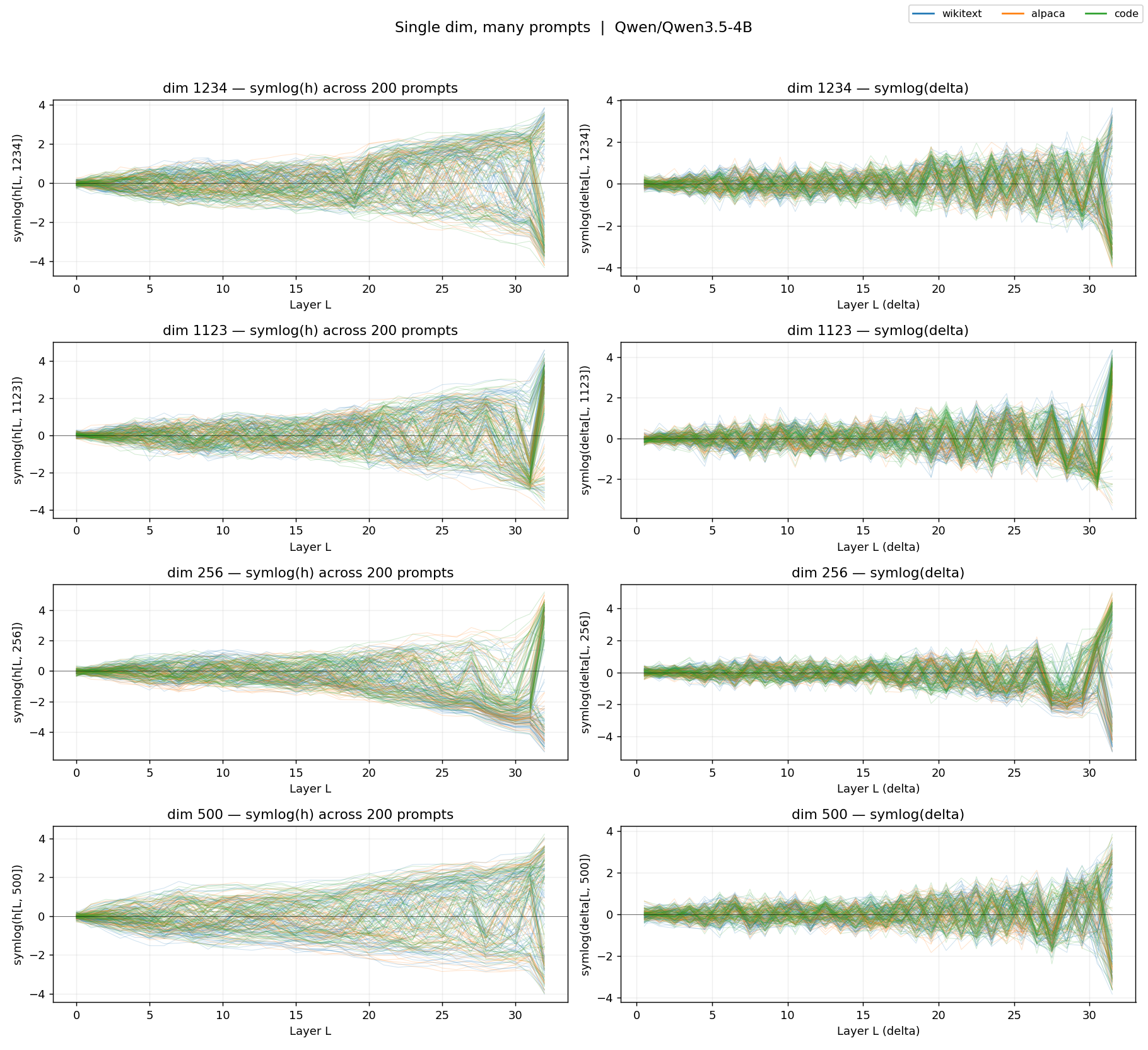}
\caption{Individual dimension trajectories (4 dims, 200 prompts colored by domain).
Each dim follows its own path; the population-level envelope emerges only when 2560
such per-dimension trajectories are overlaid.}
\label{fig:single_dim}
\end{figure}

That left the question of whether any individual dimension follows the lattice.
None of them do (Figure~\ref{fig:single_dim}). Each dimension traces its own path,
consistent within a domain but pairwise-uncorrelated with the others: mutual
information between dimension signs is negligible ($<$0.006 bits;
Appendix Table~\ref{tab:mi}), and context does not raise it appreciably. The lattice is
a population effect of overlaying many weakly-coupled channels, and no single
dimension carries it. That is what motivated us to stop reading the population and
start reading dimensions one at a time.

\section{Unsupervised Feature Discovery and SAE Comparison}
\label{app:unsupervised}

\subsection{Type-Level Head-to-Head}
\label{app:typelevel}

The first comparison is on the ground the type cache is built for: single tokens from the 175 curated
concepts, ranked against vocabulary negatives. It is pinned to the SAE's own site rather than our
discovery layer, since a released autoencoder accepts activations from the hook it was trained on and
nowhere else: \texttt{layer\_25\_width\_16k\_l0\_big} hooks \texttt{blocks.25.hook\_resid\_post},
which is $h_{26}$ here.

Both encoders read identical states. Each token is presented as $[\textsc{bos}, t]$ and read at
position 1, and the same float matrix feeds the sign side and the autoencoder; the signs derived from
it agree with the stored type cache on every dimension. Negatives are one pool of 20{,}000 vocabulary
tokens, sampled once and disjoint from every concept's anchors, and both encoders rank the same
positives against the same negatives. Anchors are split $80/20$ over three seeds, so each detector is
built from one part and scored on the other: the sign feature registers its dimensions on the build
anchors, and the autoencoder's feature is selected on the same build anchors by mean activation. Of
the 175 concepts, 153 have at least ten single-token anchors (median 11).

\begin{table}[H]
\centering
\caption{Type-level detection, Gemma 3-4B at $h_{26}$, 153 concepts $\times$ 3 seeds, anchors held
out. Both encoders score the same held-out anchors against the same 20{,}000-token negative pool from
the same states. The paired difference and its interval are over concepts.}
\label{tab:typelevel}
\begin{tabular}{lccc}
\toprule
Method & Mean AUC & Paired difference & Sign wins \\
\midrule
Sign feature (no training) & \textbf{0.990} & --- & --- \\
SAE best single feature & 0.971 & $+$0.019 $\pm$0.005 & 113/153 \\
\bottomrule
\end{tabular}
\end{table}

The sign feature is ahead by $0.019$, an interval that excludes zero, and wins on 113 of 153
concepts. It reaches this with a median of 41 registered dimensions against a null of 5, and scores
$0.995$ when the negatives are the whole vocabulary rather than the shared pool. The margin is small
because both detectors are near ceiling on single tokens; what differs is that one is read from a
frozen matrix and the other took millions of tokens and a training run.

Combining several autoencoder features through a trained classifier is the natural extension of this
comparison, and the anchor sets are too small to fit one: at a median of 11 single-token anchors, a
top-$K$ readout has about nine positives to estimate from, and its accuracy moves between $0.82$ and
$0.97$ across $K$ without ordering in $K$. We therefore report the single-feature comparison, which
needs no fitting on either side.

\subsection{Sparse Probing Against a Trained SAE}
\label{app:sparse_probing}

The comparison above is type-level: single tokens, full-vocabulary
ranking, the setting the type cache is built for. The test that puts a sign dictionary on the
trained dictionary's own ground is contextual and document-level, which is what an SAE is fitted on.
We run the sparse-probing protocol of \citet{karvonen2025saebench}: for each
class, rank dictionary features by train-set class contrast, fit a logistic probe on the top $k$, and
score held-out accuracy, with feature activations mean-pooled over the unpadded positions of each
document, identically for both dictionaries.

The dictionary on our side is the unsupervised catalog built from the \emph{unembedding matrix
alone}, re-merged to 16{,}000 concepts to match the SAE's 16{,}384. It involves no training, no
corpus, and no forward pass: the only input is a frozen weight matrix. It is also a single object,
whereas the SAE is one of thirty-four that Gemma Scope trains and releases for this model, one per
layer, because a learned dictionary's features are tied to the activations it was fitted on. We read
both at the SAE's own activation site.

We reimplement the protocol rather than importing it, and run four of its eight sparse-probing
datasets: \texttt{ag\_news}, \texttt{dbpedia\_14}, \texttt{europarl} and \texttt{bias\_in\_bios}. The
excluded four are scraped review and source-code corpora outside our approved data set, so the
numbers are not comparable to published leaderboard values. Both encoders see identical splits within
a seed, so differences are paired per (dataset, seed) over 12 units.

\begin{table}[H]
\centering
\caption{Sparse probing, Gemma 3-4B, 4 datasets $\times$ 3 seeds. ``Retained'' is BoD's share of the SAE's above-chance accuracy, $(\mathrm{BoD} - c)/(\mathrm{SAE} - c)$ for chance $c$, which is the scale on which the two are comparable; the paired difference and its 95\% interval over the 12 units are given alongside.}
\label{tab:sparse_probing}
\begin{tabular}{lccccc}
\toprule
$k$ & SAE & BoD & Retained & Difference & 95\% CI \\
\midrule
1     & 0.891 & 0.837 & 92\% & $-$0.054 & $\pm$0.028 \\
2     & 0.921 & 0.880 & 94\% & $-$0.041 & $\pm$0.019 \\
5     & 0.940 & 0.921 & 97\% & $-$0.020 & $\pm$0.008 \\
10    & 0.947 & 0.936 & 98\% & $-$0.011 & $\pm$0.005 \\
20    & 0.949 & 0.945 & 99\% & $-$0.004 & $\pm$0.003 \\
50    & 0.944 & 0.946 & 100\% & $+$0.002 & $\pm$0.006 \\
\bottomrule
\multicolumn{6}{l}{\footnotesize The sweep stops at $k = 50$. Fitting the whole dictionary answers a different question} \\
\multicolumn{6}{l}{\footnotesize than sparse probing does, and at these widths that fit is degenerate.} \\
\end{tabular}
\end{table}

A dictionary read from the unembedding matrix retains 92--100\% of the trained SAE's above-chance
accuracy across this sweep, and reaches parity at $k = 50$. Per dataset the residual gap is confined
to small $k$ on two of the four: the dictionaries are indistinguishable on
\texttt{bias\_in\_bios} at every $k$ (99.6--99.9\% retained, relative error ratio 1.01--1.04), tie
exactly on \texttt{dbpedia\_14} at $k = 20$, and differ most on \texttt{europarl} at $k \leq 2$,
which is language identification: the SAE reaches 0.999--1.000 there from $k = 2$ upward, so it
contributes a large low-$k$ gap while discriminating almost nothing above it. In no cell does BoD
retain less than 85\% of above-chance accuracy.

The direction of the small-$k$ gap is what the two methods predict. An SAE optimizes sparsity and
monosemanticity, so its best individual latents can carry a document-level class outright. Our
concepts are token-type detectors, so none of them is a ``sports article'' feature and the
corresponding information is spread across many token concepts that a document-level readout must
sum. Reading enough of them recovers it.

\textbf{One dictionary against four separately trained ones.} Gemma Scope releases a sparse
autoencoder per layer because a learned dictionary's features are tied to the activations it was
fitted on. Our dictionary has no such tie: it is read from the unembedding matrix, so there is
nothing to rebuild per layer and nothing to choose a layer for. That makes a prediction worth
testing, because it is the practical difference between the two methods. We take the same single
dictionary, unchanged, and run the same protocol at four of the released sites spanning the stack.

\begin{table}[H]
\centering
\caption{Sparse probing at four sites, 4 datasets $\times$ 3 seeds each. One BoD dictionary, read
from the unembedding matrix and never rebuilt, against a separately trained SAE at each site. Entries
are the paired BoD${}-{}$SAE accuracy difference; $^{\dagger}$ marks the cells whose 95\% interval
contains zero. Sites are named as Gemma Scope releases them with our depth index alongside.}
\label{tab:sae_layers}
\begin{tabular}{lcccccc}
\toprule
SAE site & $k{=}1$ & 2 & 5 & 10 & 20 & 50 \\
\midrule
layer 5 \ ($h_{6}$)   & $-$0.032$^{\dagger}$ & $-$0.038 & $-$0.069 & $-$0.072 & $-$0.056 & $-$0.032 \\
layer 19 ($h_{20}$)   & $-$0.002$^{\dagger}$ & $-$0.012$^{\dagger}$ & $-$0.024 & $-$0.015 & $-$0.008 & $-$0.001$^{\dagger}$ \\
layer 25 ($h_{26}$)   & $-$0.054 & $-$0.041 & $-$0.020 & $-$0.011 & $-$0.004 & $+$0.002$^{\dagger}$ \\
layer 32 ($h_{33}$)   & $-$0.073 & $-$0.041 & $-$0.026 & $-$0.012 & $-$0.008 & $+$0.003$^{\dagger}$ \\
\bottomrule
\end{tabular}
\end{table}

At three of the four sites the untrained dictionary reaches parity at $k = 50$, and at layer 19 it is
level at $k = 1$ as well (0.795 against 0.797). It also tracks the trained dictionaries' own layer
profile: where the SAE reads a class less well, so does the sign dictionary, and the two rise
together. The exception is layer 5, the only site where the gap widens rather than closes with
budget, and it is the earliest one; a dictionary read from the output matrix has least to say about
the shallowest states. Everywhere above that, one object built once matches a dictionary that was
trained for that specific layer.

\textbf{Against a sixteen-times wider dictionary.} Capacity is the other axis a trained dictionary can
be scaled along, so we repeat the comparison against the 262{,}144-feature release, sixteen times the
width of ours. On \texttt{ag\_news} and \texttt{bias\_in\_bios}
(3 seeds), the paired differences are $+0.022$, $+0.015$, $+0.008$, $+0.000$, $-0.003$ and $+0.005$
at $k = 1$ through $50$: parity at every budget, with no cell's interval excluding zero. Widening the
trained dictionary sixteen-fold does not open a gap. These two datasets are a subset of the four
above, so the numbers are comparable within this paragraph and not against
Table~\ref{tab:sparse_probing}.

\textbf{Centering, and what a single constant buys.} The features above are read as
$\mathrm{sign}(h - \mu)$, and $\mu$ has two available forms. The \emph{vocabulary} mean is the
per-dimension mean over single-token states, so it involves no text at all. The \emph{contextual}
mean is the per-dimension mean over running text: still no training and no fitted parameter, but it
does require a corpus to compute once. The two are not interchangeable, disagreeing on the sign of
48.5\% of dimensions here, because single-token states carry per-dimension offsets four times larger
than contextual states do. Reading a contextual state against a single-token mean is a frame
mismatch of the same kind as reading an intermediate state through the output unembedding.

\begin{table}[H]
\centering
\caption{Sparse probing under the two centerings, same dictionary, same splits, only $\mu$ changing (2 datasets $\times$ 3 seeds). ``wiki'' is a Wikipedia-derived contextual mean, included as a control: it shares no text with either evaluation dataset.}
\label{tab:centering}
\begin{tabular}{lccc}
\toprule
$k$ & vocabulary mean (no corpus) & contextual mean & contextual, Wikipedia \\
\midrule
1   & $-$0.059 & $-$0.018 & $-$0.014 \\
2   & $-$0.041 & $-$0.007 & $-$0.005 \\
5   & $-$0.021 & $+$0.003 & $+$0.004 \\
10  & $-$0.014 & $+$0.003 & $+$0.001 \\
20  & $-$0.004 & $+$0.004 & $+$0.003 \\
50  & $-$0.003 & $+$0.006 & $+$0.004 \\
\bottomrule
\end{tabular}
\end{table}

With the vocabulary mean the trained dictionary is ahead for $k \leq 10$ and the two are at parity
from $k = 20$. With a contextual mean they are at parity at \emph{every} budget, and the gain is
concentrated exactly where the corpus-free version was weakest. The Wikipedia column is the control
that makes this a frame correction rather than dataset overlap: it agrees with the other contextual
mean to within $0.004$ at every $k$ despite sharing no text with the evaluation sets, which is what
the $r = 0.9995$ correlation between contextual means from unrelated corpora predicts. A contextual
mean is one vector, measured once per model from arbitrary text.

We report both because they answer different questions. From frozen weights alone, with no text
anywhere in the pipeline, a dictionary matches a trained per-layer SAE at $k = 50$ on the four-dataset
sweep and from $k = 20$ on the two datasets of this table. Adding a single text-invariant constant
brings it to parity at every budget. Neither number should be quoted without its centering: the same
dictionary, splits and protocol give $-0.059$ or $-0.018$ at $k = 1$ depending only on which mean is
subtracted.

What the comparison establishes is therefore not that one dictionary is better, but what the
accuracy costs. Matching a trained per-layer dictionary to within a few points, and to parity at
$k = 50$ without a corpus or at every $k$ with one measured constant, is available from a frozen
weight matrix with no training run, no forward pass, and no separate dictionary per layer.

\subsection{Targeted Probe Perturbation}
\label{app:tpp}

Sparse probing asks whether a dictionary can \emph{read} a concept. Targeted probe perturbation asks
whether it can \emph{remove} one: ablate the features associated with a class, then check that the
class's probe degrades while the other classes' probes survive. That is selectivity, which is what a
sparsity objective is for, so it is the setting where a dense distributed code has most to lose. We
follow the protocol of \citet{karvonen2025saebench},
\[
  \mathrm{TPP}(t) \;=\; \big[\text{drop in class } t\text{'s accuracy}\big] \;-\;
                        \big[\text{mean drop in the other classes' accuracy}\big],
\]
averaged over target classes and reported per ablation budget $N$.

Ablation on the SAE side is its own reconstruction: subtract the selected latents' contribution,
$h' = h - \sum_{j} f_j(h)\,W_{\mathrm{dec}}[j]$. A sign coalition has no decoder, so we take the
analogous bounded operation: treat the selected concepts' signed patterns as a basis, fit per-state
amounts by least squares, and subtract the fit. The amount removed is therefore determined by the
state, exactly as the SAE's is, rather than by how many dimensions the coalition happens to name.

TPP is a difference of accuracy drops rather than a ratio, so it rewards whichever encoder edits the
state more. We therefore equalise the intervention: the sign dictionary's edit is rescaled per state
so that both encoders remove the same fraction of the state, $\lVert h' - h\rVert / \lVert h\rVert$,
and the comparison is between two edits of identical size.

\begin{table}[H]
\centering
\caption{Targeted probe perturbation, Gemma 3-4B at $h_{26}$, 3 datasets $\times$ 3 seeds, paired per
(dataset, seed) over 9 units. Higher is better: the intended probe breaks and the others do not. Both
encoders remove the same amount of the state, $\lVert h' - h\rVert / \lVert h\rVert$ equal to
$0.004$, $0.006$, $0.011$, $0.014$, $0.018$ and $0.024$ at the six budgets. BoD uses the
$422{,}645$-concept unsupervised catalog discovered at the SAE's own layer ($h_{26}$), with
vocabulary-mean centering, so no corpus enters this comparison. Both encoders remove the
same fraction of the state at each budget: $0.004$, $0.006$, $0.011$, $0.014$, $0.018$ and $0.024$.}
\label{tab:tpp}
\begin{tabular}{lcccc}
\toprule
$N$ ablated & BoD & SAE & Difference & 95\% CI \\
\midrule
1   & $+$0.000 & \textbf{$+$0.016} & $-$0.015 & $\pm$0.012 \\
2   & $+$0.000 & \textbf{$+$0.019} & $-$0.018 & $\pm$0.013 \\
5   & $+$0.010 & $+$0.015 & $-$0.004 & $\pm$0.007 \\
10  & $+$0.015 & $+$0.013 & $+$0.002 & $\pm$0.006 \\
20  & $+$0.022 & $+$0.023 & $-$0.001 & $\pm$0.009 \\
50  & \textbf{$+$0.044} & $+$0.027 & $+$0.018 & $\pm$0.017 \\
\bottomrule
\end{tabular}
\end{table}

The division is the same one sparse probing shows, and it runs the same way. At the smallest budgets
($N = 1, 2$) the trained dictionary is ahead by more than its interval, which is what optimizing for
sparsity buys: its single best latent is the most surgical thing available. From $N = 5$ the two are at
parity---every interval contains zero at $N = 5$, $10$ and $20$---and by $N = 50$ the sign dictionary is
ahead by more than its interval ($+0.018 \pm 0.017$). Removing a concept from a distributed code takes
several coalitions; once enough of them combine, the removal is as clean as the trained decoder's, and
cleaner still at the largest budget.

Both dictionaries are read at the SAE's layer $h_{26}$: the sign catalog is the residual-$h_{26}$
unsupervised discovery ($422{,}645$ concepts), matched to the site Gemma Scope trained its SAE at, so
the comparison isolates the encoder rather than the layer. Matching the layer matters: a catalog
discovered at the unembedding and applied at $h_{26}$ trails by roughly $0.01$--$0.03$ at every budget,
because ablation is sensitive to reading a concept in the frame it will be removed in. Reading wants a
different granularity: in \S\ref{app:sparse_probing} the merged 16{,}000-concept dictionary is at parity
across the sweep where the full catalog trails by $0.010$ to $0.020$ at $k = 5$ to $20$. The two tasks
want different granularities from the same discovery procedure.

This places targeted ablation with the read and write results of \S\ref{sec:write_catalog}. A sign
coalition is an address: it is read at 0.81--0.95 AUC in context (\S\ref{sec:contextual}), and
subtracting one in its own coordinates is close to inert, which is why the read catalog suppresses a
concept but does not induce one. Ablation is a write-side operation on a read-side object.

\subsection{Unsupervised Feature Quality}

The unsupervised catalogs are validated without human labels by asking whether each concept detects its own held-out members against random vocabulary negatives. Held-out AUC is 0.985--1.000 across all eight catalogs, with mean between-concept member overlap at or below 0.001 (Table~\ref{tab:unsupervised}).

\section{Probe and MLP Ablation}
\label{app:probe}

To test whether cross-dimension structure provides practical benefit, we train a 2-layer MLP (128 hidden neurons, ReLU) that can learn arbitrary nonlinear dimension combinations. Training data scaled from 50 to 2000 examples per concept. 80/20 train/test split, test AUC reported.

\begin{table}[H]
\centering
\caption{Probe ablation: does cross-dim structure exist at any training scale? (Qwen 3.5-4B, 175 concepts, test AUC).}
\label{tab:probe_ablation}
\resizebox{\textwidth}{!}{%
\begin{tabular}{lcccc}
\toprule
Train size (pos+neg) & Mean LogReg AUC & Mean MLP AUC & MLP wins ($\Delta > 0.01$) & Mean gap \\
\midrule
100 (50+50) & 0.9964 & 0.9931 & 4/175 (2\%) & $-$0.003 \\
400 (200+200) & 0.9977 & 0.9957 & 0/175 (0\%) & $-$0.002 \\
1000 (500+500) & 0.9991 & 0.9980 & 0/175 (0\%) & $-$0.001 \\
2000 (1000+1000) & 0.9992 & 0.9984 & 0/175 (0\%) & $-$0.001 \\
4000 (2000+2000) & 0.9992 & 0.9987 & 0/175 (0\%) & $-$0.001 \\
\bottomrule
\end{tabular}}
\end{table}

The MLP never outperforms LogReg on any concept at any training scale. The gap is consistently negative: additional capacity is a liability, not an asset. Since LogReg converges to axis-aligned weights (no learned rotation) and the MLP with full cross-dim capacity adds nothing, cross-dimension structure provides no practical benefit beyond what per-dim reading already captures.

\textbf{Hard-negative evaluation.} Full-vocabulary AUC could be inflated by class imbalance. We test against semantically adjacent concepts as hard negatives (animals vs food/body parts, weapons vs vehicles). Hard-negative AUC remains 0.92--1.0 (mean 0.977).

\section{Full Concept Data}
\label{app:concepts}

The 175 concepts span:

\begin{table}[H]
\centering
\caption{Feature strength tiers (Qwen 3.5-4B, $h_{24}$).}
\label{tab:tiers}
\begin{tabular}{llc}
\toprule
Tier & Representative concepts & Count \\
\midrule
Strong ($\geq 0.85$) & physics, country, body, language, electronics & 14 \\
Solid (0.80--0.85) & biology, math, metal, emotion, code, conjunction & 72 \\
Moderate (0.75--0.80) & weather, religion, fruit, tool, weapon, currency & 75 \\
Weak (0.70--0.75) & music\_genre, greeting, narrative, sadness, bird & 14 \\
\bottomrule
\end{tabular}
\end{table}

Strong features are ontologically discrete (physics terms are fundamentally unlike non-physics terms). Weak features are fine-grained subtypes (``bird'' $\subset$ animal), gradient/degree concepts (``sadness'' as a degree of emotion), or context-dependent (``greeting'' depends on pragmatic use). All 175 exceed null calibration.

\section{Layer Sweep and Per-Head Specialization}
\label{app:layers}

\begin{figure}[h]
\centering
\includegraphics[width=0.75\textwidth]{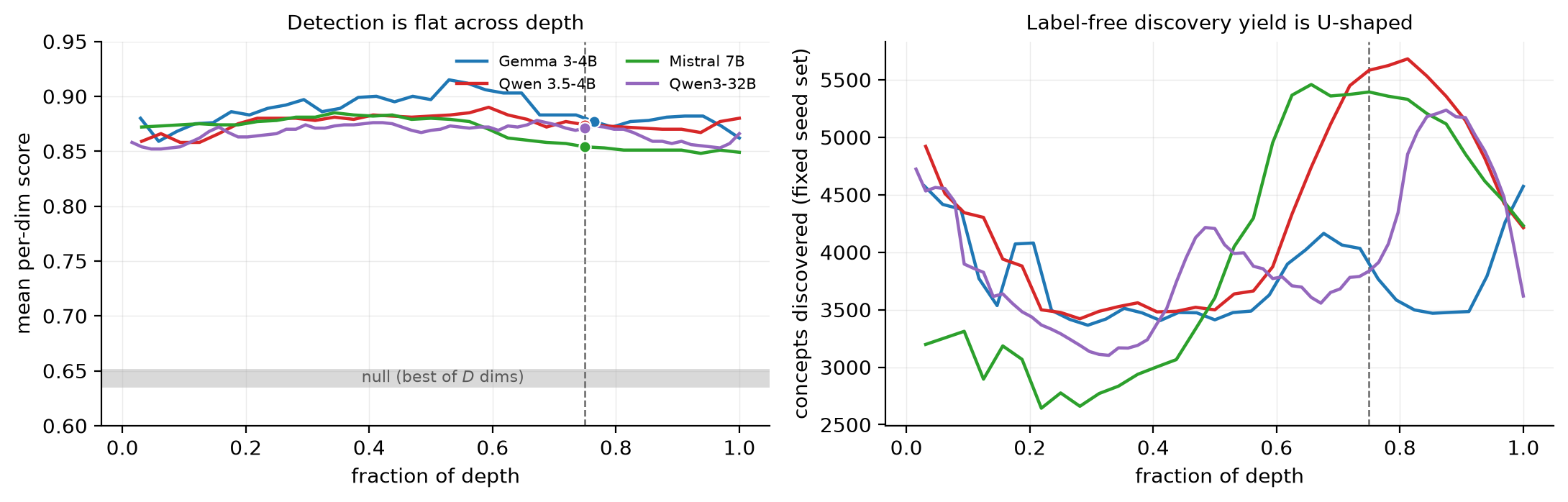}
\caption{Every layer of all four models, 175 curated concepts (171 for Mistral), \textsc{bos}
caches. \textbf{Left:} mean per-dimension score against fraction of depth. All four models stay inside
a narrow band far above the null (grey, the best-of-$D$ chance level from 100 random anchor sets);
dots mark the read layer used elsewhere in the paper (dashed line, 75\% of depth). \textbf{Right:}
label-free discovery yield, the number of concepts discovered from the same fixed set of seed tokens at
each layer. Unlike detection it varies with depth: on all four models it is U-shaped, with a trough at
22--33\% of depth, and the 75\% read layer sits past the trough.}
\label{fig:layer_sweep}
\end{figure}

\begin{table}[H]
\centering
\caption{Detection across depth, all four models, 175 curated concepts (171 for Mistral) with
full-vocabulary negatives. Layers are indexed by fraction of depth so the models align. Each cell
gives the layer, mean per-dimension score, and concepts discoverable at $\kappa_{\text{reg}} = 0.75$. Every layer
of every model is read at 0.85--0.92, and the whole-stack span is 0.026--0.056.}
\label{tab:layer_sweep}
\resizebox{\textwidth}{!}{%
\begin{tabular}{lcccc}
\toprule
Depth & Gemma 3-4B (34) & Qwen 3.5-4B (32) & Mistral 7B (32) & Qwen3-32B (64) \\
\midrule
$0.03$ & $h_{1}$\; 0.880\; 175/175 & $h_{1}$\; 0.859\; 174/175 & $h_{1}$\; 0.872\; 171/171 & $h_{2}$\; 0.854\; 174/175 \\
$0.25$ & $h_{8}$\; 0.889\; 175/175 & $h_{8}$\; 0.880\; 175/175 & $h_{8}$\; 0.878\; 171/171 & $h_{16}$\; 0.866\; 175/175 \\
$0.50$ & $h_{17}$\; 0.897\; 175/175 & $h_{16}$\; 0.882\; 175/175 & $h_{16}$\; 0.880\; 171/171 & $h_{32}$\; 0.869\; 175/175 \\
$0.75$ & $h_{26}$\; 0.877\; 175/175 & $h_{24}$\; 0.874\; 175/175 & $h_{24}$\; 0.854\; 168/171 & $h_{48}$\; 0.871\; 175/175 \\
$1.00$ & $h_{34}$\; 0.862\; 175/175 & $h_{32}$\; 0.880\; 175/175 & $h_{32}$\; 0.849\; 169/171 & $h_{64}$\; 0.866\; 175/175 \\
\midrule
whole-stack span & 0.056 & 0.032 & 0.037 & 0.026 \\
score peak & $h_{18}$ & $h_{19}$ & $h_{11}$ & $h_{43}$ \\
min.\ discoverable & 99\% & 99\% & 98\% & 99\% \\
\bottomrule
\end{tabular}}
\end{table}

This appendix gives the per-layer detail behind \S\ref{sec:layer_invariance}. Layers are indexed by
fraction of depth so that models of 32, 34 and 64 layers share an axis. The right panel is label-free: the
same fixed set of seed tokens is fractured on every layer's residual sign cache with identical
settings, so any change in the number of concepts found is a property of the layer, not of seed
variance. Yield is high near the embedding, falls to a trough at 22--33\% of depth on all four models,
and recovers in the later layers; the fixed 75\% read layer sits past the trough.

\section{Random-Init Control}
\label{app:trained_random}

To verify that discovered features require training, we run identical discovery procedures on randomly initialized models (same architecture, random weights, seed=42).

\begin{table}[H]
\centering
\caption{Random-init control: feature discovery on untrained models.}
\label{tab:random_init}
\begin{tabular}{lccccc}
\toprule
Model & Random score & Trained score & Discoverable ($\geq$0.75) & Unsupervised yield \\
\midrule
Qwen 3.5-4B & 0.608 & 0.801 & 0/175 & 0/200 \\
Gemma 3-4B & 0.603 & 0.763 & 0/175 & 0/200 \\
Mistral 7B & 0.679 & 0.844 & 11/175 & 1/200 (incoherent) \\
\bottomrule
\end{tabular}
\end{table}

Random-init models produce no monosemantic, semantically coherent features. The residual AUC above 0.50 reflects input embedding similarity that passes through random projections (stronger for Mistral's smaller 32K vocabulary). The one Mistral ``feature'' (5 dims, tokens: ``widely'', ``itched'', ``installed'', ``EqualTo'', ``adding'') is semantically incoherent and fires on 3.4\% of vocabulary (not monosemantic).

This is the feature-level counterpart of the population-level trained-vs-random
comparison in Figure~\ref{fig:trained_vs_random} (Appendix~\ref{app:lattice}): the
expanding envelope survives random initialization, but the internal structure that
makes individual dimensions readable does not.

\section{Toy Superposition Model}
\label{app:superposition}

To compare the toy regime against real transformers (\S\ref{sec:discussion}), we replicate Elhage et al.'s toy autoencoder ($x \mapsto W^\top W x$ with ReLU, bottleneck $d \ll n$, sparsity $s$) and test whether per-dim sign patterns recover the ``superposed'' features.

\begin{table}[H]
\centering
\caption{Toy superposition replication (Elhage's model, $\mathrm{ReLU}(W^\top h + b)$, seed 42). \textbf{Superposition} is established on Elhage's own criteria rather than asserted: $\sum_i D_i$ is the number of dimensions in use, where $D_i = \lVert W_i\rVert^2 / \sum_j (\hat{W_i}\!\cdot\!W_j)^2$ is feature $i$'s share of a dimension, and it saturates its bound $d$ while more than $d$ features are represented, so features must share dimensions; feat./dim is $1/D_i$, the mean number of features per dimension; interf.\ is the mean $|\cos|$ between represented feature columns. \textbf{Decoding} compares per-dim sign agreement, which never sees $W$, against projection onto $w_i/\lVert w_i\rVert$ --- the model's own decoder, not a baseline we introduce. Sign matches projection on every configuration. Relaxing the bottleneck from $d{=}5$ to $d{=}20$ drops interference roughly five-fold and features-per-dimension from 3.2--4.1 to 2.2--2.5, the regime effect of \S\ref{sec:discussion}.}
\label{tab:toy_superposition}
\resizebox{\textwidth}{!}{%
\begin{tabular}{lcccccccc}
\toprule
 & & & \multicolumn{3}{c}{Superposition (Elhage's criteria)} & \multicolumn{2}{c}{Decoding} \\
\cmidrule(lr){4-6}\cmidrule(lr){7-8}
Config & $n$ & $d$ & Repr. & $\sum_i D_i$ & feat./dim & interf.\ $|\cos|$ & sign & projection \\
\midrule
$s{=}0.05$ & 20 & 5 & 16 & 5.00 & 3.2 & 0.333 & 100\% (16/16) & 100\% \\
$s{=}0.02$ & 40 & 5 & 16 & 5.00 & 3.2 & 0.333 & 100\% (16/16) & 100\% \\
$s{=}0.01$ & 100 & 5 & 17 & 4.99 & 4.1 & 0.345 & 88\% (15/17) & 88\% \\
$s{=}0.05$ & 100 & 20 & 41 & 19.92 & 2.2 & 0.052 & 100\% (41/41) & 100\% \\
$s{=}0.02$ & 200 & 20 & 46 & 19.88 & 2.5 & 0.068 & 100\% (46/46) & 100\% \\
\bottomrule
\end{tabular}}
\end{table}

Superposition is present on Elhage's criteria: $\sum_i D_i$ saturates $d$ while 16--17 features are represented in $d{=}5$ and 41--46 in $d{=}20$, so roughly 2--4 features share each dimension, and the mean $|\cos|$ between represented columns is $0.33$ at $d{=}5$. Non-orthogonality alone would not show this, since $n > d$ vectors cannot be mutually orthogonal; the shared-dimension count is what carries it. In that regime per-dim sign agreement detects 88--100\% of represented features and matches projection onto each feature's own direction on every configuration, so where geometric interference is real and substantial, reading signs loses nothing against the model's own decoder.

This shows sign matching is a sufficient decoder of the toy's features. To characterize what the hidden state itself stores, we measure two further properties below.

\textbf{Sign-vs-magnitude and cross-dim coupling in the toy.} The detection result above shows sign matching identifies which feature is active, but does not characterize how the toy's hidden state stores content. We measure two further properties: sign-only reconstruction quality (forcing $|h_d|$ to the row-mean magnitude and passing through $W^\top$), and pairwise mutual information between dim signs in $h$ (Eq.~\ref{eq:mi}, same procedure as Table~\ref{tab:mi}).

\begin{table}[H]
\centering
\caption{Toy hidden-state analysis (Elhage's model, seed 42): reconstruction cost of discarding magnitude, and pairwise MI between dim signs. Both variants are decoded through the \emph{model's own} decoder, $\mathrm{ReLU}(W^\top\!\cdot + b)$; sign-only rescales each row of $\mathrm{sign}(h)$ to its mean $|h|$ first, so what is measured is the loss of magnitude \emph{information}, not of scale. Both are computed over the represented features. $R^2$ is the fraction of importance-weighted variance explained, and ``retained'' is the sign-only $R^2$ as a share of the full $R^2$; we report these rather than the sign-to-full error ratio, which is ill-conditioned because its denominator approaches zero as the model fits those features. Real-transformer cross-dim MI for comparison (Table~\ref{tab:mi}, mean columns): 0.0005--0.0036 bits, which the $d{=}20$ rows here already reach.}
\label{tab:toy_sign_encoding}
\resizebox{\textwidth}{!}{%
\begin{tabular}{lccccccc}
\toprule
 & \multicolumn{2}{c}{Weighted MSE} & \multicolumn{3}{c}{Variance explained ($R^2$)} & \multicolumn{2}{c}{Sign MI (bits)} \\
\cmidrule(lr){2-3}\cmidrule(lr){4-6}\cmidrule(lr){7-8}
Config & full & sign-only & full & sign-only & retained & mean & max \\
\midrule
$n{=}20$, $d{=}5$, $s{=}0.05$    & 0.00353 & 0.00990 & 0.856 & 0.597 & 70\% & 0.0708 & 0.094 \\
$n{=}40$, $d{=}5$, $s{=}0.02$    & 0.00098 & 0.00389 & 0.905 & 0.625 & 69\% & 0.0520 & 0.130 \\
$n{=}100$, $d{=}5$, $s{=}0.01$   & 0.00045 & 0.00196 & 0.902 & 0.570 & 63\% & 0.0427 & 0.091 \\
$n{=}100$, $d{=}20$, $s{=}0.05$  & 0.00028 & 0.00501 & 0.975 & 0.546 & 56\% & \textbf{0.0063} & 0.045 \\
$n{=}200$, $d{=}20$, $s{=}0.02$  & 0.00006 & 0.00217 & 0.983 & 0.431 & 44\% & \textbf{0.0056} & 0.053 \\
\bottomrule
\end{tabular}}
\end{table}

Two patterns emerge. First, magnitude carries real information about \emph{how much}, and signs alone still carry most of it: sign-only reconstruction retains 63--70\% of the variance the full state explains at $d{=}5$ and 44--56\% at $d{=}20$, reaching $R^2 = 0.43$--$0.63$ in absolute terms with magnitudes discarded entirely. This is the complement of the detection result in Table~\ref{tab:toy_superposition}, not a tension with it, and together they state the division of labour directly: the sign pattern suffices to identify \emph{which} features are active (88--100\%, matching the model's own decoder), while magnitudes refine \emph{how much} of each. A code in which signs alone reconstructed perfectly would place no information in magnitude at all, which is not what we claim. We report $R^2$ rather than the sign-to-full error ratio because that ratio is ill-conditioned here: its denominator shrinks toward zero as the model fits the represented features, so it swings by more than an order of magnitude under changes that leave the model untouched, whereas $R^2$ is normalised by the data and bounded above by 1.

Second, cross-dim MI distinguishes the regimes, and this is the load-bearing observation. The toy bottleneck at $d{=}5$ shows mean pairwise MI of 0.043--0.071 bits, an order of magnitude above real transformers (Table~\ref{tab:mi}: 0.0005--0.0036 bits). As the bottleneck relaxes to $d{=}20$, toy MI falls to 0.0056--0.0063 bits, inside real-transformer range, and the interference measured on $W$ falls with it, from $|\cos| = 0.33$ to $0.05$--$0.07$ (Table~\ref{tab:toy_superposition}). Features-per-dimension drops from 3.2--4.1 to 2.2--2.5 over the same change, and the spread of $1/D_i$ widens from 2.9--4.1 to 1.0--8.4, so with room available the model gives some features an almost dedicated dimension instead of packing them all. Every coupling signature that motivates a learned rotation weakens together as the artificial pressure comes off. The geometric coupling that defines superposition is a function of bottleneck pressure, not a fixed property of the architecture.

\section{Cross-concept Polysemy: Case Detail}
\label{app:polysemy}

The 77 cross-concept polysemy cases (\S\ref{sec:contextual}) each pair a polysemous word with a concept-sense sentence and an other-sense sentence; we score the target word's contextual representation against the word's concept feature in both. ``Correct'' means the concept-sense context scores higher. The target token is verified to be located in both sentences; cases failing this check are excluded (2 on Qwen, varies by tokenizer). We report all cases without selection.

\begin{table}[H]
\centering
\caption{Representative cross-concept cases (Qwen 3.5-4B, $\tau{=}0.70$). Score = fraction of feature dims whose contextual sign matches the concept polarity.}
\label{tab:polysemy_cases}
\begin{tabular}{llccc}
\toprule
Word & Concept & Concept-sense & Other-sense & $\Delta$ \\
\midrule
train & vehicle & 0.931 & 0.448 & $+$0.483 \\
jeep & vehicle\_land & 0.783 & 0.217 & $+$0.565 \\
owl & bird & 0.938 & 0.562 & $+$0.375 \\
pine & tree & 0.935 & 0.613 & $+$0.323 \\
date & fruit & 0.875 & 0.583 & $+$0.292 \\
seal & animal & 0.750 & 0.525 & $+$0.225 \\
\midrule
\multicolumn{5}{l}{\emph{Representative failures (reported, not excluded):}} \\
copper & metal & 0.780 & 0.923 & $-$0.143 \\
calf & body & 0.628 & 0.752 & $-$0.124 \\
whip & weapon & 0.667 & 0.889 & $-$0.222 \\
\bottomrule
\end{tabular}
\end{table}

Strong, ontologically distinct pairs (vehicle, tree, bird) separate cleanly; failures concentrate in concepts whose senses are both concrete and adjacent (body parts, metals). The remaining cases follow the same pattern.

\textbf{Threshold robustness.} The effect is not a tuned $\tau$: accuracy is stable across $\tau{\in}[0.55,0.70]$ and increases as the feature broadens (lower $\tau$, more dimensions voting).

\section{Causal Sign-Flip: Per-concept Detail}
\label{app:causal_flip}

This appendix provides the per-concept numbers behind the causal intervention summarized in \S\ref{sec:contextual}. For each model and concept we build the feature's sign pattern from the type cache, then during a live forward pass flip the feature's signs at all positions and report the change in the mean logit of the concept's target tokens (greedy decoding). Patch layers: Qwen 3.5-4B $h_{24}$, Gemma 3-4B $h_{34}$, Mistral 7B $h_{24}$, Qwen3-32B $h_{60}$ (the near-output layer required by the depth-dependence of \S\ref{sec:discussion}); the Gemma flip is applied at block 33, whose FFN writes the $h_{34}$ residual used for detection (block $l{-}1$ writes $h_l$).

\begin{table}[H]
\centering
\caption{Per-concept target-logit change under sign-flip (full coalition, $\tau{=}0.6$). \textbf{Away}: signs flipped away from expected. \textbf{Toward}: same dimensions and magnitudes forced to the expected sign (isolates sign from magnitude). \textbf{Random}: equal number of random dimensions. Away suppresses on every concept that responds; toward and random do not.}
\label{tab:causal_per_cat}
\begin{tabular}{llccc}
\toprule
Model & concept & Away & Toward & Random \\
\midrule
\multirow{5}{*}{Qwen 3.5-4B} & animals & $-$11.55 & $+$0.81 & $-$2.24 \\
 & numbers & $-$13.90 & $-$0.42 & $-$2.93 \\
 & colors & $-$11.81 & $-$2.68 & $-$3.24 \\
 & food & $-$10.10 & $+$0.73 & $+$0.03 \\
 & countries & $-$10.22 & $-$1.17 & $-$0.98 \\
\midrule
\multirow{5}{*}{Gemma 3-4B} & animals & $-$20.32 & $+$2.07 & $-$3.59 \\
 & numbers & $-$19.11 & $+$0.73 & $-$3.31 \\
 & colors & $-$24.27 & $+$2.12 & $-$4.12 \\
 & food & $-$20.71 & $+$1.73 & $-$1.97 \\
 & countries & $-$24.46 & $+$1.92 & $-$4.32 \\
\midrule
\multirow{5}{*}{Mistral 7B} & animals$^\ddagger$ & $-$1.86 & $-$0.70 & $-$1.19 \\
 & numbers & $-$7.16 & $-$1.64 & $-$2.27 \\
 & colors & $-$6.99 & $-$0.74 & $-$1.13 \\
 & food & $-$5.10 & $-$0.21 & $-$1.25 \\
 & countries & $-$6.23 & $-$0.68 & $-$3.94 \\
\midrule
\multirow{5}{*}{Qwen3-32B} & animals & $-$5.10 & $+$0.29 & $-$0.94 \\
 & numbers & $-$8.98 & $-$2.31 & $-$2.14 \\
 & colors & $-$7.38 & $-$1.84 & $-$0.17 \\
 & food & $-$6.12 & $-$0.30 & $-$0.03 \\
 & countries$^\ddagger$ & $+$1.20 & $-$0.97 & $-$2.15 \\
\bottomrule
\multicolumn{5}{l}{\footnotesize $^\ddagger$Each model has one weak concept that does not respond (animals/Mistral,} \\
\multicolumn{5}{l}{\footnotesize countries/Qwen3-32B); 4/5 concepts respond strongly on every model.} \\
\end{tabular}
\end{table}

\textbf{Concept-specificity via the disjoint control.} A natural concern is that flipping any large trained sign-coalition might degrade the model's general next-token competence rather than suppressing the specific concept. It does not. Because features share dimensions ($\sim$35\% overlap), flipping a \emph{different} concept's coalition partially flips the target's; the apparent cross-concept damage is entirely this overlap. Removing the shared dimensions (flipping only the part of another concept's coalition that is disjoint from the target) leaves the target essentially untouched, indistinguishable from a random flip of equal size.

\begin{table}[H]
\centering
\caption{Disjoint specificity control (target-logit change on concept $A$). \textbf{A (full)}: flip $A$'s own coalition. \textbf{Disjoint}: flip another concept's coalition with $A{\cap}B$ removed. \textbf{Random}: equal-size random flip. Disjoint $\approx$ random: the causal effect is concept-specific, not general damage.}
\label{tab:disjoint}
\begin{tabular}{llccc}
\toprule
Model & Concept $A$ & A (full) & Disjoint & Random \\
\midrule
\multirow{3}{*}{Qwen 3.5-4B} & animals & $-$11.55 & $-$0.34 & $-$0.65 \\
 & numbers & $-$13.90 & $-$0.60 & $-$2.39 \\
 & colors & $-$11.81 & $-$0.17 & $-$3.18 \\
\midrule
\multirow{3}{*}{Gemma 3-4B} & animals & $-$20.32 & $-$3.47 & $-$3.16 \\
 & colors & $-$24.27 & $-$1.45 & $-$2.27 \\
 & countries & $-$24.46 & $-$1.77 & $-$1.37 \\
\midrule
\multirow{3}{*}{Mistral 7B} & numbers & $-$7.16 & $-$0.25 & $+$0.04 \\
 & colors & $-$6.99 & $-$0.41 & $-$0.21 \\
 & food & $-$5.10 & $-$0.13 & $-$0.27 \\
\midrule
\multirow{3}{*}{Qwen3-32B} & numbers & $-$8.98 & $-$0.11 & $-$0.45 \\
 & colors & $-$7.38 & $-$0.77 & $+$0.06 \\
 & food & $-$6.12 & $-$0.02 & $-$2.08 \\
\bottomrule
\end{tabular}
\end{table}

Across all four models, flipping a disjoint coalition produces an effect statistically indistinguishable from a random flip, while flipping the concept's own coalition suppresses it by $5$--$24$ logits. The causal effect is therefore specific to the feature, and the coalition, not any single dimension, is the unit at which it acts.

\section{Write Catalog: Steering Detail}
\label{app:write_catalog}

This appendix reports the per-concept steering results behind Table~\ref{tab:control} and example generations. All runs use the write target $\boldsymbol{\tau}_c$ (\S\ref{sec:method_write}) injected as a sign-agreement coalition into the attention output ($W_O$) input under closed-loop presence control, with the sign-correct write and a logit-clean decode (no repetition penalty, no $n$-gram blocking). Each model is scored over 48 trials (12 concepts $\times$ 4 held-out prompts) by an LLM judge against the model's own unsteered continuation; a run counts as a success only if the continuation both refers to the concept and is at least as fluent as the baseline. The judge is greedy and deterministic, so a single pass suffices.

\paragraph{Per-concept success.} Table~\ref{tab:steer_perconcept} gives the closed-loop success for each of the twelve concepts, and the residual failures track the presence--coherence frontier rather than broken targets. \emph{Body} is the hardest concept ($7/16$ across the four models, $1/4$ on both Gemma and Qwen 3.5-4B): its part-word senses surface only at a push that also breaks the sentence. \emph{Weapon} is next ($9/16$), failing entirely on Qwen3-32B ($0/4$) yet carried $4/4$ by both base models. Qwen 3.5-4B is the softest overall at $29/48$, under-inducing at its coherent operating point. No concept fails everywhere: \emph{food} is the easiest at $15/16$, reaching $4/4$ on three of the four models, and \emph{vehicle} is $4/4$ on three (the two base models and Qwen3-32B).

\begin{table}[H]
\centering
\caption{Per-concept closed-loop steering success (of 4 held-out prompts). Sign-correct write, attention pathway, per-model settings of Table~\ref{tab:control}.}
\label{tab:steer_perconcept}
\begin{tabular}{lcccc}
\toprule
Concept & Gemma 3-4B & Mistral 7B & Qwen 3.5-4B & Qwen3-32B \\
\midrule
animal  & 3 & 3 & 2 & 3 \\
number  & 4 & 2 & 4 & 2 \\
color   & 4 & 1 & 3 & 4 \\
food    & 3 & 4 & 4 & 4 \\
country & 3 & 3 & 1 & 4 \\
emotion & 2 & 3 & 4 & 4 \\
body    & 1 & 2 & 1 & 3 \\
fruit   & 3 & 4 & 3 & 2 \\
vehicle & 4 & 4 & 1 & 4 \\
metal   & 3 & 4 & 3 & 2 \\
weapon  & 4 & 4 & 1 & 0 \\
tech    & 3 & 3 & 2 & 2 \\
\midrule
Total   & 37/48 (77\%) & 37/48 (77\%) & 29/48 (60\%) & 34/48 (71\%) \\
\bottomrule
\end{tabular}
\end{table}

\paragraph{Example generations.} To show that the write target, not the prompt, drives the output, we hold the prompt fixed and vary only the concept: the same context is steered to different concepts (attention pathway, closed-loop, logit-clean greedy decode). The shared prompt is shown in roman and the steered continuation in \emph{italic}. These are successful trials from the same run as Table~\ref{tab:steer_perconcept}, which reports every concept including the failures.

\smallskip
\noindent\textbf{Gemma 3-4B} --- ``Yesterday I went to the \dots''
\begin{itemize}\itemsep2pt
\item (animal) \emph{zoo with my friends. We saw a dog, a cat, a bird, a monkey, a tiger, a lion, a giraffe}
\item (food) \emph{local supermarket for dinner. I was craving a good meal and I was not disappointed. I had a delicious chicken and rice}
\item (fruit) \emph{local market and bought some fresh strawberries. I was so excited to get them home and make a strawberry shortcake}
\item (weapon) \emph{range with my new 1911. I had a 100 round box of 230 grain FMJ. I shot}
\end{itemize}

\noindent\textbf{Mistral 7B} --- ``After a long week, he decided to \dots''
\begin{itemize}\itemsep2pt
\item (animal) \emph{take his dog, a dog named Bella, for a walk. He was walking along the beach with his dog when he saw a}
\item (food) \emph{go to a restaurant to have dinner. He was tired and hungry. He was looking forward to a good meal.}
\item (country) \emph{take a trip to the country of his birth, the United States of America. He was born in the United States,}
\item (body) \emph{take a shower and wash his body. He was so tired that he fell asleep in the shower. He woke up to the}
\end{itemize}

\noindent\textbf{Qwen 3.5-4B} --- ``She opened the door and saw \dots''
\begin{itemize}\itemsep2pt
\item (animal) \emph{the animals, and the animals remained still, as if they were waiting for her to speak, but she had}
\item (color) \emph{the colors of the morning, a brilliant, shifting tapestry of gold and violet that seemed to pulse}
\item (food) \emph{the food table set for three, with a single, unlit candle flickering in the center of the table}
\item (fruit) \emph{the fruit of her labors, though the city had not yet fully claimed the land, and the air still held}
\end{itemize}

\noindent\textbf{Qwen3-32B} --- ``Yesterday I went to the \dots''
\begin{itemize}\itemsep2pt
\item (animal) \emph{market and saw a dog. The dog was small, with a coat that shimmered in the sunlight like a patchwork}
\item (color) \emph{market, and the colors were a riot of spring---vivid greens, warm reds, and soft yellows spilled across}
\item (food) \emph{market and bought a chicken, a bag of rice, and a bottle of olive oil. The sun was high and the street}
\item (country) \emph{city and walked through the streets, past the buildings, and into the country where the land stretched}
\end{itemize}

\section{Axis-Aligned Signs in FFN Activations}
\label{app:ffn_signs}

The read catalog lives in the residual stream, and steering acts through the attention pathway (\S\ref{sec:write_catalog}). This appendix reports a structural observation adjacent to those results: the same axis-aligned sign patterns are also present one step earlier, in the FFN that feeds the residual, in the neuron \emph{activations}. These are correlational activation-space measurements; we do not establish that these neurons are the causal writer of a concept, and we do not use them for detection or control anywhere in the paper.

\textbf{Neuron activations are themselves sign-readable.} The neuron \emph{activations} (the SwiGLU intermediate $\mathrm{SiLU}(\text{gate}(x))\cdot\text{up}(x)$) form a readable per-dim sign space. This is not automatic: the activation sign is $\mathrm{sign}(\text{gate}(x))\cdot\mathrm{sign}(\text{up}(x))$, a nonlinear product of two projections, so axis-aligned structure could be destroyed by the gate. We hook block $l{-}1$, read each neuron's activation sign for the single-token vocabulary, and run the identical per-dim score and feature procedure used for the residual stream (same 175 concepts, same 20K-token negative sample, same $\tau$ and null calibration).

\begin{table}[H]
\centering
\caption{Per-dim sign features in FFN \emph{neuron activations} vs.\ the residual stream, under an identical harness (175 concepts, 171 for Mistral; full-vocabulary negatives; $\tau{=}0.70$). The neuron read hooks block $l{-}1$, whose output writes $h_l$.}
\label{tab:neuron_bod}
\begin{tabular}{lcccccc}
\toprule
& & \multicolumn{2}{c}{Per-dim max AUC} & \multicolumn{2}{c}{Feature AUC} & \\
\cmidrule(lr){3-4}\cmidrule(lr){5-6}
Model & Neurons & Neuron & Residual & Neuron & Residual & Exceed null \\
\midrule
Qwen 3.5-4B ($h_{24}$) &  9,216 & 0.819 & 0.801 & 0.990 & 0.983 & 175/175 \\
Gemma 3-4B ($h_{34}$)  & 10,240 & 0.818 & 0.773 & 0.976 & 0.968 & 174/175 \\
Mistral 7B ($h_{24}$)  & 14,336 & 0.861 & 0.844 & 0.994 & 0.993 & 171/171 \\
Qwen3-32B ($h_{48}$)   & 25,600 & 0.787 & 0.792 & 0.973 & 0.980 & 165/175 \\
\bottomrule
\end{tabular}
\end{table}

Despite the nonlinear gate, neuron activations match the residual stream within $\pm0.05$ on per-dim score and within $\pm0.02$ at the feature level, exceeding null calibration on 165--175 of 175 concepts, with no random anchor set forming a feature (null $p_{99}=0.500$). The Bag-of-Dims sign structure is therefore not specific to the residual stream; it is already present, axis-aligned and sign-readable, in the FFN's own activation space and survives the SwiGLU gate.

\section{Extended Discussion}
\label{app:discussion}

\textbf{Why are dimensions readable one at a time?} We observe this as an emergent property of training, not an architectural guarantee. It is consistent with the negligible pairwise MI between dimension signs (Table~\ref{tab:mi}) and with the structure surviving through learned projections. We emphasize that this is \emph{functional} independence (cross-dimension capacity adds no practical value for feature reading), not a claim of strict statistical independence.

\textbf{Combinatorial coding.} We discover tens to hundreds of thousands of features per model (\S\ref{sec:feature_discovery}) from 2560--5120 dimensions, each using a median of 15--106 dims. Features remain independently detectable, with between-concept member overlap at or below 0.001: scoring uses sign agreement on each feature's own registered dim subset, not a global operation, so two features can share dims without interfering. The MI measurements ($<$0.006 bits; Table~\ref{tab:mi}) confirm that shared dims introduce no practical coupling. The capacity is combinatorial rather than geometric: features need not compete for orthogonal directions.

\textbf{Relationship to the superposition hypothesis.} The superposition hypothesis \citep{elhage2022toy} proposes that transformers encode more features than dimensions by placing features at non-orthogonal directions that geometrically interfere, motivating learned rotations (SAEs) for decoding. Our results suggest the encoder/decoder distinction matters here: features are encoded as sign-magnitude pairs (sign carrying content, magnitude carrying strength), and per-dim sign matching is a sufficient decoder regardless of whether the encoder packs features at non-orthogonal angles.

With $D$ binary dimensions, the number of distinct sign-pattern features (subsets of $k$ dims with specified signs) is:
\begin{equation}
    \text{Capacity} = \sum_{k=1}^{D} \binom{D}{k} \cdot 2^k = 3^D - 1
\end{equation}
For $D = 2560$, this yields $\sim 10^{1220}$ possible features, effectively infinite. The address space is large enough that geometric packing is not the only available mechanism for representing many features.

We replicate Elhage et al.'s toy autoencoder ($x \mapsto W^\top W x$ with bottleneck $d \ll n$; full results in Appendix~\ref{app:superposition}). Superposition emerges as reported: $\sum_i D_i$ saturates the bottleneck while 16--46 features are represented, so 2--4 features share each dimension. 88--100\% of the represented features are recoverable from per-dim sign patterns alone (AUC $> 0.7$), and on every configuration this matches projection onto the feature's own direction, which is the model's own decoder; in a 20-dim bottleneck encoding 41 features, sign-based detection recovers all 41. Sign-only reconstruction (Appendix~\ref{app:superposition}, Table~\ref{tab:toy_sign_encoding}) retains 63--70\% of the explained variance at $d{=}5$ with magnitudes discarded, the complement of that detection result: signs identify which features are active, magnitudes refine how much of each, the same division of labour we find in real LMs (Table~\ref{tab:sign_prediction}).

We do not refute superposition: at $d{=}5$ the toy shares each dimension between 3--4 features with mean $|\cos| = 0.33$ and cross-dim sign MI of 0.043--0.071 bits (Tables~\ref{tab:toy_superposition}--\ref{tab:toy_sign_encoding}), real properties of its bottleneck regime. We add an empirical observation: the same toy already exhibits sign-as-content / magnitude-as-strength, the structure we identify in real transformers, and per-dim sign matching is a sufficient decoder even where geometric interference is present. Real transformers do not show the toy's coupling signatures: pairwise MI is $<$ 0.006 bits (Table~\ref{tab:mi}, vs.\ 0.043--0.071 in the toy bottleneck), the cross-dim MLP buys nothing over per-dim reading, and probes converge to axis-aligned weights. As the toy's bottleneck relaxes, its MI drops into real-transformer range (0.0056--0.0063 bits at $d{=}20$) and its interference falls five-fold. The geometric coupling that defines superposition appears as a function of bottleneck pressure, and the regime where that pressure binds does not appear to be the regime real transformers operate in.

\textbf{Contextual updating.} The flipping dimensions are low-magnitude dims below the feature threshold. The cross-concept polysemy test (\S\ref{sec:contextual}) shows context modulates the feature read: a word scores higher against its concept feature in the matching context. So these context-dependent dims carry sense information rather than noise; whether they also encode syntactic role or reference is open. The high-magnitude, feature-registered dims preserve type identity, which is why type-level features generalize to context: they read from the stable high-magnitude partition. Contextual pairwise MI is lower than type-level, consistent with context diversifying rather than entangling dimensions.

\textbf{The causal intervention is depth-gated, and read and write sites differ.} Write-side control (\S\ref{sec:write_catalog}) is effective across a \emph{band} of layers, not a single one: injecting the write target from a fraction of depth up to a per-model endpoint forms the concept's consensus as it builds, rather than overwriting a formed one late. The coherent band endpoint is model-specific ($0.62$--$0.88$ of depth) and, notably, is \emph{not} the last layer: pushing the write toward the output raises concept presence but over-drives the surrounding text into repetition, so the usable endpoint sits below the output where presence and fluency trade off best (\S\ref{sec:write_catalog}). The read-side sign-flip shows a complementary depth structure that is diagnostic rather than a capability: it shifts the target logits only when applied near the output (on the 64-layer Qwen3-32B, absent at the $h_{48}$ read layer but present at $h_{60}$, despite near-identical detection AUC, 0.91 vs 0.90). A sign pattern corrupted deep in the stack is re-derived by the intervening layers from earlier context before it reaches the logits, so a feature is \emph{readable} across a wide band but logit-effective only near the output. This re-derivation explains why early single-layer perturbation studies found sign flips inert: the corruption is corrected downstream.

\textbf{Induction requires a separate object from detection.} The read-side sign-flip is asymmetric: flipping a feature's read signs \emph{away} lowers its target logits, but forcing them \emph{toward} expected does not induce the concept, because the high-magnitude read dimensions already carry the expected sign. Induction is not a property of the read dimensions at all; it lives on a different set. This is what motivates the write target (\S\ref{sec:write_catalog}): the dimensions it favors are nearly disjoint from those whose sign reads the concept (rank correlation $0.01$--$0.11$), and a coalition selected by agreement with $\boldsymbol{\tau}_c$, injected through the attention output pathway under closed-loop control (\S\ref{sec:write_catalog}), induces the concept in generation where perturbing the read signs cannot. The asymmetry thus resolves into a \emph{where}, not a \emph{what}: suppression and induction operate on separate dimensions, not via a magnitude-vs-sign distinction on the same ones. The sign account holds throughout: reading is per-dimension sign agreement, and the write target is likewise a per-dimension sign (of the output-projection consensus), not a learned direction.

\textbf{Modality universality.} The appearance of per-dim sign structure in DINOv2 and AST, trained on images and audio respectively with no shared data or vocabulary, suggests the structure is driven by the optimization process (gradient descent on transformer architectures) rather than properties of natural language specifically. The DINOv2 vs.\ ViT-Base comparison is particularly informative: same architecture, same evaluation data, but self-supervised vs.\ supervised training. Both exhibit per-dim sign structure, with classification merely sharpening it (11/12 vs.\ 9/12 superclasses, more dims per concept above threshold). This pattern, structure present without supervision and amplified by supervision, implies that per-dim specialization is a convergent property of residual transformers trained by gradient descent, regardless of objective.

\textbf{Why do different objectives converge?} Why next-token prediction, self-distillation, and classification all produce the same per-dim structure is an open question. We note only that the shared factor is the architecture: a residual stream with additive FFN writes and sparse activation. A formal information-theoretic treatment is beyond the scope of this work.

\textbf{Limitations.} (1) Scale: tested on 1.4--32B language models and base-sized (86M parameter) vision/audio models; whether the structure persists at 70B+ is open, though consistent results from 1.4B to 32B suggest no degradation. (2) The two catalogs are complementary: the read-side sign-flip suppresses (it cannot induce, because induction lives on disjoint dimensions), while the write target induces via a coalition injected through the attention pathway under closed-loop control (\S\ref{sec:write_catalog}). Both are concept-specific. We validate suppression at the logit level (Table~\ref{tab:causal_flip}, mean target-logit change) and do not claim it yields fluent suppressed \emph{generation}. Removing a concept that the surrounding context keeps re-deriving is harder than inducing one, and fluent suppression is future work. Write-side control is set per model and shared across all twelve concepts (the write band, basin depth, and coalition size), not tuned per concept, so where a concept resists at that setting we report the failure rather than adjusting it; these sit at the presence-coherence frontier (\S\ref{sec:write_catalog}). Induction succeeds for 60--77\% of trials across four models, with base models at the top of that range and the two reasoning-tuned models more variable. The near-ceiling detection AUC (0.998--1.000) also suggests a non-interventional use: real-time monitoring of per-dim sign patterns for safety-relevant features, enabling early stopping or fallback. (3) Type-level features read type identity from the stable high-magnitude dims; systematically cataloging the contextual-role features encoded by the flipping 42\% is future work. (4) Per-dim score varies only slightly across architectures (0.854--0.877) and we do not have an account of what drives the remaining spread. (5) The 175 curated language model concepts are a small subset of the model's feature inventory: unsupervised discovery yields 37{,}000--682{,}000 concepts per catalog (\S\ref{sec:feature_discovery}), and since the count is set by a merge threshold we do not claim it is the total. (6) Both catalogs now port to non-language transformers: detection and K/V discovery on all three, and the disjoint write target with attention-pathway output control on the two with classifier heads (ViT-Base, AST; \S\ref{sec:cross_modality}). Two gaps remain. The write target requires an output head to read $\boldsymbol{\tau}_c$ from, so DINOv2 (self-supervised, no classifier) is excluded from the control result; deriving a write target for a headless model is future work. And the cross-modal control is validated at the level of the output decision (capture rate, specificity, and the sign-shuffled/amplifier controls of Table~\ref{tab:cross_control}); because these models are non-generative there is no fluency axis to evaluate, so the coherence question that the closed-loop release addresses in language (\S\ref{sec:write_catalog}) does not arise here.

\section{Catalog Construction and Control: Full Method}
\label{app:method}

\subsection{Single-Token Type Cache}

Feature discovery requires no contextual data. Type-level semantics, what a token \emph{is}, are already encoded in the sign pattern after a single forward pass of that token alone. No other token is present, so self-attention has no cross-token content to mix in and the resulting hidden state reflects the token's identity as processed by the full transformer stack, without influence from surrounding words.

We construct a \textbf{type-level cache} by running every vocabulary token through the model on its own:

\begin{enumerate}
    \item For each of the $V$ tokens in the vocabulary ($\sim$248K for Qwen 3.5-4B, $\sim$262K for Gemma, $\sim$32K for Mistral, $\sim$152K for Qwen3-32B):
    \begin{itemize}
        \item Feed the sequence $[\textsc{bos}, t]$ and read position 1, the token's own position
        \item Extract the hidden state at target layers
        \item Store the $D$-dimensional state
    \end{itemize}
    \item Result: a matrix $\mathbf{H} \in \mathbb{R}^{V \times D}$ per layer.
\end{enumerate}

\textbf{The beginning-of-sequence prefix.} Every sequence the model saw in training began with \textsc{bos}, so a bare token is off-distribution input and the stack processes it accordingly. Prefixing \textsc{bos} restores the trained format at the cost of nothing that matters here: \textsc{bos} is a content-free delimiter, identical for every token in the vocabulary, so it contributes no token-specific information and cannot make one token look more like a given concept than another. The read position still holds exactly one word's worth of meaning. The cache is therefore context-free in the sense the method requires, and in-distribution in the sense the model requires. The prefix is not doing the work, only removing an obstacle to it: ablating it on Gemma drops discovery from 175 of 175 concepts to 137, and introduces a pronounced mid-stack dip that is absent with the prefix in place (Appendix~\ref{app:layers}).

\textbf{Layers.} The builder caches every layer in one pass, so nothing below is specific to a
choice of depth; \S\ref{sec:layer_invariance} shows detection and decoding work at all of them. The
catalogs and tables that follow are reported in two sign spaces: the embedding matrix, and the
residual stream at a fixed three-quarters of depth for every model ($h_{26}$ for Gemma of 34,
$h_{24}$ for Qwen 3.5-4B and Mistral of 32, $h_{48}$ for Qwen3-32B of 64). The choice is a
convention, not a tuned optimum: detection is flat across the stack
(\S\ref{sec:layer_invariance}), so we fix one standard fraction and report it uniformly; 75\%
sits past the trough of the U-shaped label-free discovery yield (22--33\% of depth) and below the
output layer (Appendix~\ref{app:layers}). Where a comparison requires a particular site, such as
the layer at which a released SAE was trained, we state it there.

The cache requires one forward pass per vocabulary token, batched, and is computed once per model. Cost scales with vocabulary size rather than parameter count: a single layer of Mistral's 32K vocabulary takes under a minute on one GPU, and the 152K-token, 64-layer Qwen3-32B takes about 11 minutes sharded over four. Since all subsequent analysis operates only on sign patterns, the cache can be stored as packed bits (1 bit per dimension), reducing storage 32$\times$ relative to float32: 93 MB for the full Qwen3-32B vocabulary at one layer. All feature discovery, feature building, and evaluation in this paper operate on this cache, requiring no sentences, no prompts, and no gradient computation.

\begin{figure}[t]
\centering
\includegraphics[width=\textwidth]{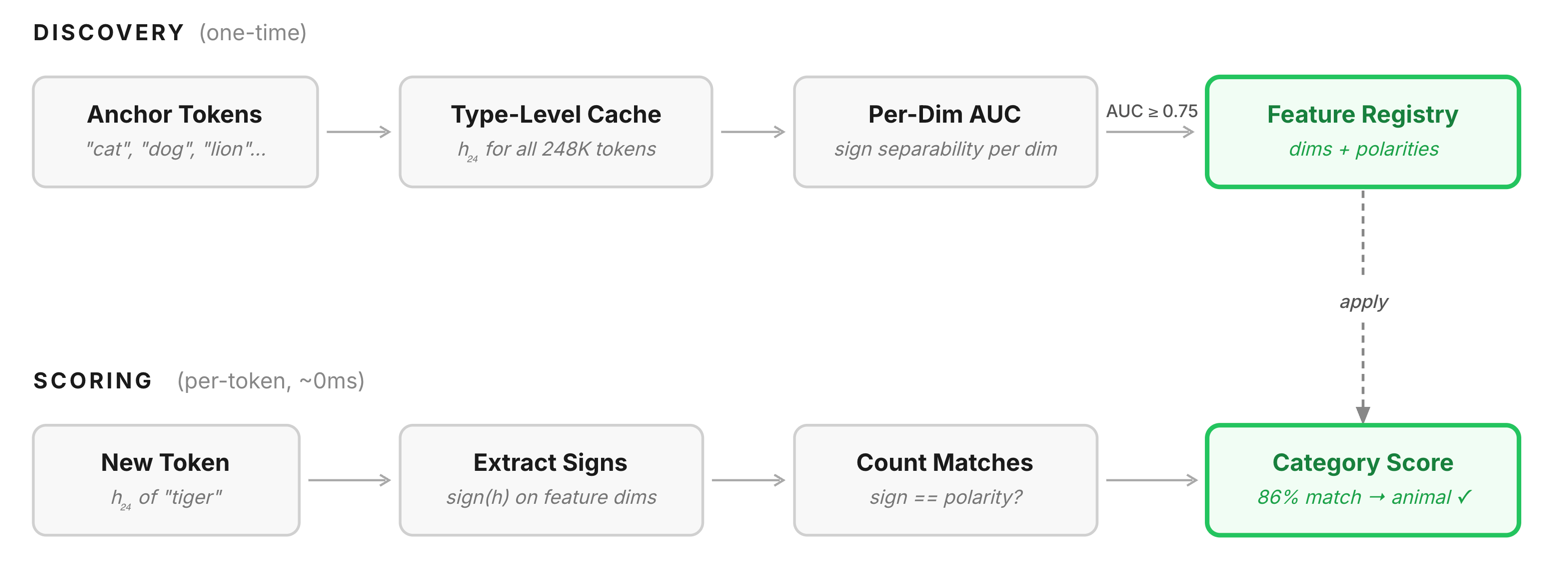}
\caption{The Bag-of-Dims discovery and scoring pipeline. Top: one-time discovery builds a feature registry from anchor tokens and the type-level cache. Bottom: scoring any new token is instant via sign matching on registered dims.}
\label{fig:pipeline}
\end{figure}

\subsection{Unsupervised Discovery}

The procedure above needs anchor tokens, and anchor tokens need labels. The catalogs that form this
paper's main result (\S\ref{sec:feature_discovery}) use none: the only input is the sign cache
itself.

\textbf{Mean-relative signs.} For unlabeled search we take signs relative to each dimension's
vocabulary mean, $\mathrm{sign}(h_d - \mu_d)$, rather than relative to zero. With curated anchors the
per-dimension score already absorbs each dimension's baseline firing rate, but an unlabeled
neighbor search has no such reference, so a dimension that is positive for most of the vocabulary
would otherwise dominate every Hamming distance. This recentering is diagonal, one scalar per
dimension: no rotation and no cross-dimension mixing. It changes the sign of \emph{only} the
baseline-biased dimensions. Raw and mean-relative signs agree on 95\% of near-balanced dimensions
and differ on the strongly biased ones, with per-dimension disagreement tracking baseline bias at
$\rho = 0.99$. The dimensions that discriminate a concept keep their raw standard-basis sign;
centering only stops uninformative broadcast dimensions from swamping the distance.

\textbf{One seed yields several concepts.} We take each vocabulary token in turn as a seed, collect
its 60 nearest neighbors by sign agreement, and then \emph{fracture} that neighborhood into clusters.
Fracture works by consensus. For each dimension we measure the fraction of the neighborhood that
shares the seed's sign: dimensions with consensus at or above $0.80$ are the seed's \emph{concept
core}, shared across the whole neighborhood, while dimensions with consensus at or below $0.60$ are
the \emph{split dimensions}, where the neighborhood divides. We then cluster the neighbors by their
sign pattern on the split dimensions alone (agglomerative, Hamming), with the number of clusters
chosen by a parameter-free minimum-description-length criterion (Occam, at most 10 clusters). This
step matters: a token's sign-neighborhood is a superposition of every sense and grouping the token
participates in, so fracturing demultiplexes it and a single seed contributes several distinct
concepts rather than one blurred average. Each cluster's member set defines a candidate concept, and
clusters with fewer than 8 members are discarded.

\textbf{A parameter-free acceptance test.} A candidate is accepted on a compression criterion rather
than a tuned threshold. We measure the cluster's excess bit-deficit against a matched
random-content null of the same size and shape, and require at least one bit per member. A real
coalition compresses its members, in that their sign patterns are predictable from membership; an
accidental grouping of the same size does not. Nothing here is fit to the data.

\textbf{Feature.} Accepted clusters are reduced to the same $(\mathcal{T}_c,
\boldsymbol{\pi}_c)$ form as \S\ref{sec:discovery}, so a concept discovered without labels is read
by exactly the same sign-agreement count as a curated one.

\textbf{Merging and filtering.} Concepts discovered from different seeds that describe the same thing
are merged when their member sets overlap at Jaccard $\geq \tau_{\text{merge}} = 0.90$; an identical
$(\mathcal{T}_c, \boldsymbol{\pi}_c)$ is dropped outright. We merge on member sets rather than sign
agreement because sign-agreement merges collapse into the residual stream's global bias field,
leaving no clean threshold, whereas member-set Jaccard is bias-free: unrelated concept pairs sit at
$\approx 0.004$ mean overlap, far below any usable threshold. Setting $\tau_{\text{merge}}$ high, at
$0.90$, therefore collapses only near-duplicate coalitions and keeps genuinely distinct concepts
apart, which is why the catalogs are large and their between-concept overlap is near zero
(Table~\ref{tab:unsupervised}). $\tau_{\text{merge}}$ is the resolution knob referred to throughout:
raising it keeps finer distinctions and yields more concepts, lowering it merges more aggressively
into fewer, coarser ones. Two final filters remove what cannot be a concept: candidates whose members
are majority dead vocabulary (untrained or reserved tokens, which cluster tightly because they are
untrained rather than because they mean anything), and coalitions of fewer than 8 dimensions, which
cannot specify a token set.

\textbf{Quality metrics.} We report two, neither of which uses labels. \emph{Held-out AUC} splits
each concept's members, builds the feature from one half, and asks whether it detects the held-out
half against random vocabulary negatives: does the concept detect its own members? \emph{Redundancy}
is the mean member-set Jaccard between concepts: are they distinct, or restatements of each other?

\subsection{Extension to Attention and MLP Subspaces}

Nothing in the procedure is specific to the residual stream: it needs a sign per dimension, and any
tensor in the forward pass supplies one. We read three further subspaces at the same layer:
\begin{itemize}
    \item $\text{sign}\!\big(\text{RoPE}(\text{Norm}(\mathbf{h}) \cdot W_k^\top)\big)$ for K dimensions
    \item $\text{sign}(\text{Norm}(\mathbf{h}) \cdot W_v^\top)$ for V dimensions
    \item $\text{sign}\!\big(\sigma(\text{Norm}(\mathbf{h}) \cdot W_{\mathrm{gate}}^\top) \odot (\text{Norm}(\mathbf{h}) \cdot W_{\mathrm{up}}^\top)\big)$ for the MLP gate activation
\end{itemize}
where $\text{Norm}$ is the model's pre-block normalization (RMSNorm for all language models here),
RoPE is the rotary position embedding applied to K after projection, $\sigma$ is the gating
nonlinearity (SiLU), and $\odot$ is elementwise. The gate activation is what the MLP's down
projection consumes, so it is read at the feed-forward width $d_{\mathrm{ffn}}$, which is 2.4--5$\times$
wider than the residual stream and therefore offers correspondingly finer resolution.

We build the caches by capturing these tensors through the full compute path (normalization,
projection, RoPE for K, gating for the MLP) rather than reconstructing them from weights, then apply
the identical discovery procedure and compare concept-level AUC against the residual-stream
baseline at the same layer (\S\ref{sec:attention}). Where a model's attention departs from the
standard per-layer form, K and V are not well defined and we say so rather than reporting a number.

\subsection{Steering control: coalition, closed loop, and judge}

\paragraph{Control: a write coalition in the attention output pathway.} The write target $\boldsymbol{\tau}_c$ lives in output (residual) space. To use it for control we inject not into the residual directly but into the \emph{input} of the attention output projection $W_O$ (\texttt{o\_proj}), the per-head attention-output space $z$ that $W_O$ maps additively into the residual. Attention's contribution is linear in $z$ (no gating nonlinearity), so a sign-aligned write in this space passes to the residual intact. At each write layer $\ell$ we score every coordinate $j$ of the $W_O$ input by how well its weight column's signs agree with the target:
\begin{equation}
    s^{\ell}_c[j] \;=\; \frac{1}{D}\sum_{d=1}^{D} \mathrm{sign}\!\big(W_O^{\ell}[d,j]\big)\,\boldsymbol{\tau}_c[d],
    \label{eq:coalition_score}
\end{equation}
and take the coalition $\mathcal{C}^{\ell}_c$ of the top-$K$ coordinates by $|s^{\ell}_c[j]|$, with polarity $\rho[j]=\mathrm{sign}(s^{\ell}_c[j])$. Before selecting, we drop the low-\emph{spread} coordinates: those whose agreement score is near-identical across all concepts, a shared ``generic amplifier.'' This keeps the coalition concept-specific. Then, during generation, at the last position, we push only the coordinates whose current sign disagrees with the target, and only up to a margin:
\begin{equation}
    z^{\ell}[j] \;\leftarrow\; z^{\ell}[j] \;+\; g(t)\,\max\!\big(0,\; m^{\ell} - z^{\ell}[j]\,\rho[j]\big)\,\rho[j], \qquad j \in \mathcal{C}^{\ell}_c,
    \label{eq:steer}
\end{equation}
where the target margin $m^{\ell} = \beta\,Q_{0.999}(|z^{\ell}|)$ is a fraction $\beta$ of the layer's own $99.9$th-percentile activation magnitude, so the write is scale-free across layers and models. A coordinate already past the margin on the correct side has zero deficit and is left untouched; only the disagreeing coordinates are moved, and only far enough to clear the margin. Leaving the already-correct coordinates free to keep churning is what prevents the collapse into a repeated concept token: $\beta$ sets the basin \emph{depth}, not a blunt additive force.

\paragraph{Closed-loop presence control.} Rather than push with a fixed gain, we release the write
once the concept is present, and we read presence where it is legible earliest: in the attention
\emph{value} states, not the residual. At a sensor layer near the start of the write band (band
start $+1$), we register the concept's sign coalition in the value subspace exactly as elsewhere
(\S\ref{sec:discovery}, the same $\kappa_{\text{reg}}$ on a value-space sign cache), and after each
generated token we read its raw agreement $p(t)$, the fraction of that coalition's value-state
dimensions whose sign matches the concept's polarity. Raw value-state agreement is a far stronger
presence signal than a residual read at the write layers (AUC $0.82$ against near-chance), which is
why the loop is closed in value space. The write runs at full strength for a warm-up of $F$ tokens
so the concept can establish, then releases: the first time $p(t)$ clears a release threshold $T$,
the gain drops to zero and the write stops for the remainder of the generation, with a fallback cap
forcing release if the threshold is never met. $T$ is one of the per-model settings (it does not
transfer across models) and is applied to every concept alike.

The decode itself stays clean: we decode greedily (temperature $0$) with \emph{no} repetition penalty
and \emph{no} $n$-gram blocking, and cap generation at 30 new tokens, so the text comes from the
intervention and not from a decode-time heuristic.

\paragraph{Hyperparameters.} The write spans a band of layers rather than a single layer, from a
fraction of depth up to a per-model endpoint. Injecting across the band forms the concept's
consensus as it builds, instead of overwriting a formed one late. Five settings are chosen per model
and then shared across all twelve concepts: the layer band, the basin depth $\beta$, the coalition
size $K$, the value-space sensor layer, and the release threshold $T$
(\S\ref{sec:write_catalog}, Table~\ref{tab:control}). Three more take a single value on every model: the
spread cutoff ($0.5$), the warm-up floor $F$ ($6$ tokens), and the
generation length ($30$ tokens). The two reasoning-tuned models are additionally run through their
chat template, with their thinking tokens banned during decode. None of these is trained. They are decode-time settings, while the feature basis and the
write target come from the weights alone.

The gain is unregulated for the first $F$ generated tokens (the warm-up floor), after which the
release trigger takes over. The write therefore establishes the concept at full strength and
the loop then releases it, rather than pushing throughout.

Spread is computed as the standard deviation of a coordinate's agreement score across the concepts
being steered, so the coalition is defined relative to that set; we report the concept set used with
each model.

\textbf{Suppression} remains a read-side operation, unchanged from \S\ref{sec:contextual}. Flipping the concept's \emph{read} signs away from $\boldsymbol{\tau}_c$ at every generated token removes it, the complement of the write-side induction here.

\paragraph{Scoring steering with an LLM judge.} A lexical hit-test has two blind spots. It cannot recognize that a suppressed ``animal'' surfacing as ``panda'' still evokes the concept, and it cannot tell fluent prose from a grammatical-but-derailed fragment. Both fixes would be hidden tuning: a hand-curated word list, or a coherence threshold.

We instead score each generated continuation with an LLM judge (a strong instruction-tuned model, greedy-decoded). It returns two judgments. \emph{Concept presence} asks whether the continuation meaningfully refers to the target concept, judged by meaning rather than exact words. \emph{Coherence} asks whether the continuation is at least as fluent as the model's own unsteered continuation of the same prompt. The judge sees the unsteered baseline alongside the steered text, so coherence measures the steering \emph{delta}: an intervention is penalized only if it degrades fluency the base model already had. A trial counts as a success only if the continuation is both concept-present and coherent.

The judge is greedy (temperature zero) and deterministic, so re-scoring a fixed transcript reproduces the same counts; we report a single pass of 48 trials (12 concepts $\times$ 4 prompts) per model. It scores only the open language models under study, and it is an evaluation instrument, not part of the training-free steering method.


\section{Additional Experiment Tables}
\label{app:evtables}
Full versions of the per-model tables summarized in Section~\ref{sec:experiments}.

\begin{table}[t]
\centering
\footnotesize
\caption{Prefill toxicity detection on RealToxicityPrompts (test-split AUC). A
gradient-free count over the sign bits tracks a trained linear probe to within a
few points on all eight models; both fit the same labeled split. The early column
is read at 12\% depth, except Pythia-1.4B at 17\% (0.78 at 12\%, i.e.\ its third
block); Qwen3-32B is read to 38\% depth. The training-free, label-free gate is
reported in the text.}
\label{tab:guardrail}
\begin{tabular}{@{}l c c c@{}}
\hline
Model & sign count @ 12--17\% depth & sign count (peak) & trained probe (ceiling) \\
\hline
Mistral-7B & 0.87 & 0.92 & 0.95 \\
Gemma-3-4B & 0.80 & 0.92 & 0.94 \\
Qwen3.5-4B & 0.82 & 0.90 & 0.94 \\
Qwen3-32B  & 0.89 & 0.92 & 0.94 \\
\hline
Mamba-2.8B  & 0.83 & 0.92 & 0.91 \\
OLMoE-1B-7B & 0.80 & 0.91 & 0.94 \\
Pythia-1.4B & 0.88 & 0.93 & 0.94 \\
RWKV-4-3B   & 0.81 & 0.92 & 0.94 \\
\hline
\end{tabular}
\end{table}

\begin{table}[t]
\centering
\footnotesize
\caption{Correctness monitor on the factual battery. The margin of the late-layer
sign-vote readout (top candidate minus runner-up, pure sign agreement) is a
calibrated confidence: it separates correct from incorrect readouts (monitor
AUC), and keeping the most-confident half of prompts raises accuracy from the
base rate to 0.85 to 1.00. No training.}
\label{tab:monitor}
\begin{tabular}{@{}l c c c c@{}}
\hline
Model & monitor depth & readout acc & monitor AUC & acc on top-half margin \\
\hline
Mistral-7B & 81\% & 0.69 & 0.92 & 0.98 \\
Gemma-3-4B & 76\% & 0.72 & 0.86 & 0.92 \\
Qwen3.5-4B & 81\% & 0.80 & 0.91 & 1.00 \\
Qwen3-32B  & 94\% & 0.75 & 0.73 & 0.85 \\
\hline
Mamba-2.8B  & 88\% & 0.73 & 0.98 & 1.00 \\
OLMoE-1B-7B & 88\% & 0.69 & 0.91 & 0.96 \\
Pythia-1.4B & 75\% & 0.68 & 0.91 & 0.96 \\
RWKV-4-3B   & 94\% & 0.67 & 0.89 & 0.94 \\
\hline
\end{tabular}
\end{table}

Pairwise mutual information between dimension signs $S_i, S_j$:
\begin{equation}
    I(S_i; S_j) = \sum_{a,b \in \{0,1\}} P(S_i{=}a, S_j{=}b) \log_2 \frac{P(S_i{=}a, S_j{=}b)}{P(S_i{=}a)\, P(S_j{=}b)}
    \label{eq:mi}
\end{equation}

\begin{table}[H]
\centering
\caption{Sign-only next-token prediction accuracy ($N=200$ prompts per model).}
\label{tab:sign_prediction}
\begin{tabular}{llcccc}
\toprule
Model & Method & Top-1 & Top-5 & Top-10 & Top-100 \\
\midrule
\multirow{4}{*}{Qwen 3.5-4B} & Full $\mathbf{h} \cdot W$ & 100\% & 100\% & 100\% & 100\% \\
 & $\text{sign}(\mathbf{h}) \cdot W$ & 52\% & 80\% & 86.5\% & 96\% \\
 & Top-800 loud (sign only) & 64\% & 89\% & 95\% & 99\% \\
 & Random sign permutation & 0\% & 0\% & 0\% & 0\% \\
\midrule
\multirow{4}{*}{Gemma 3-4B} & Full $\mathbf{h} \cdot W$ & 100\% & 100\% & 100\% & 100\% \\
 & $\text{sign}(\mathbf{h}) \cdot W$ & 42\% & 67\% & 76.5\% & 96\% \\
 & Top-800 loud (sign only) & 48.5\% & 79\% & 89\% & 100\% \\
 & Random sign permutation & 0\% & 0\% & 0\% & 0\% \\
\midrule
\multirow{4}{*}{Mistral 7B} & Full $\mathbf{h} \cdot W$ & 100\% & 100\% & 100\% & 100\% \\
 & $\text{sign}(\mathbf{h}) \cdot W$ & 65\% & 93.5\% & 96.5\% & 100\% \\
 & Top-800 loud (sign only) & 75.5\% & 93\% & 97.5\% & 100\% \\
 & Random sign permutation & 0\% & 0\% & 0\% & 0\% \\
\midrule
\multirow{4}{*}{Qwen3-32B} & Full $\mathbf{h} \cdot W$ & 100\% & 100\% & 100\% & 100\% \\
 & $\text{sign}(\mathbf{h}) \cdot W$ & 36.5\% & 55.5\% & 64.5\% & 96.5\% \\
 & Top-800 loud (sign only) & 50.5\% & 75.5\% & 83.5\% & 98\% \\
 & Random sign permutation & 0\% & 0\% & 0\% & 0.5\% \\
\bottomrule
\end{tabular}
\end{table}

\begin{table}[H]
\centering
\caption{Pairwise mutual information between dimension signs (1000 random pairs per condition).}
\label{tab:mi}
\begin{tabular}{lcccc}
\toprule
Model & Type-level mean & Type-level max & Contextual mean & Contextual max \\
\midrule
Qwen 3.5-4B & 0.0008 & 0.016 & 0.0009 & 0.024 \\
Gemma 3-4B & 0.0003 & 0.015 & 0.0009 & 0.014 \\
Mistral 7B & 0.0012 & 0.020 & 0.0011 & 0.011 \\
Qwen3-32B & 0.0036 & 0.073 & 0.0005 & 0.009 \\
\bottomrule
\multicolumn{5}{l}{\footnotesize 1{,}000 random dimension pairs over 25{,}600 tokens; contextual is 200 texts $\times$ 128 positions.} \\
\end{tabular}
\end{table}

\begin{table}[H]
\centering
\caption{Probe weight analysis: the probe learns axis-aligned voting.}
\label{tab:probe}
\begin{tabular}{p{0.72\textwidth}c}
\toprule
Metric & Value \\
\midrule
Sign agreement, all registered dims (AUC $> 0.70$; 306--493 dims per concept) & 91--98\% \\
Sign agreement, top 40 dims per concept by AUC & 96--99\% \\
Spearman $\rho$ (signed weight vs.\ signed per-dim score, registered dims) & 0.77--0.82 \\
MLP (2-layer, 128 hidden) vs.\ sign-counting advantage, all four models & $+0.0006$ to $+0.0053$ \\
\bottomrule
\end{tabular}
\end{table}

\begin{table}[H]
\centering
\caption{Cross-concept polysemy: fraction of cases where the concept feature scores the concept-sense context higher than the other-sense context, and pooled AUC, at the $\tau$ that maximizes AUC per model (full sweep stable across $\tau{\in}[0.55,0.70]$). On-thesis (per-dim sign feature scoring, no whole-vector similarity), token-validated, all cases reported.}
\label{tab:polysemy}
\begin{tabular}{lccc}
\toprule
Model & Cases & Accuracy & Mean score margin \\
\midrule
Qwen 3.5-4B  ($h_{24}$) & 56/69 & \textbf{81\%} & $+$0.044 \\
Qwen3-32B    ($h_{48}$) & 57/69 & 83\% & $+$0.026 \\
Mistral 7B   ($h_{24}$) & 33/40 & 82\% & $+$0.025 \\
Gemma 3-4B   ($h_{26}$) & 48/70 & 69\% & $+$0.015 \\
\bottomrule
\end{tabular}
\end{table}

\begin{table}[H]
\centering
\caption{Causal sign-flip during the live forward pass (mean target-logit change, 5 concepts per model). \textbf{Away}: flip the feature's signs away from expected. \textbf{Toward}: force the same dims to their expected sign (same dims, same magnitudes; isolates sign from magnitude). \textbf{Disjoint}: flip a different feature's coalition with the target's shared dims removed (isolates concept-specificity from overlap). \textbf{Random}: flip an equal number of random dimensions.}
\label{tab:causal_flip}
\begin{tabular}{lcccc}
\toprule
Model & Away & Toward & Disjoint & Random \\
\midrule
Qwen 3.5-4B ($h_{24}$) & $-$10 to $-$14 & $\approx 0$ & $\approx 0$ & $\approx 0$ \\
Gemma 3-4B ($h_{34}$)  & $-$19 to $-$24 & $\approx 0$ & $\approx 0$ & $\approx 0$ \\
Mistral 7B ($h_{24}$)  & $-$5 to $-$7   & $\approx 0$ & $\approx 0$ & $\approx 0$ \\
Qwen3-32B ($h_{60}$)$^\dagger$ & $-$5 to $-$9 & $\approx 0$ & $\approx 0$ & $\approx 0$ \\
\bottomrule
\multicolumn{5}{l}{\footnotesize $^\dagger$The intervention requires patching near the output; on the 64-layer} \\
\multicolumn{5}{l}{\footnotesize model the effect appears at $h_{60}$, not the $h_{48}$ read layer (\S\ref{sec:discussion}).} \\
\end{tabular}
\end{table}

\begin{table}[H]
\centering
\caption{The write catalog across four models (12 concepts). \textbf{Disjoint}: rank correlation between per-dim read strength and write-target polarity, and mean read-AUC of top write-target dimensions (0.5 = chance). \textbf{Specificity}: cross-concept write-target Jaccard. Control results are in \S\ref{sec:write_catalog}.}
\label{tab:write_catalog}
\begin{tabular}{lcccc}
\toprule
 & Gemma 3-4B & Mistral 7B & Qwen 3.5-4B & Qwen3-32B \\
\midrule
read--write $\rho$ (disjoint)        & 0.11 & 0.01 & 0.02 & 0.01 \\
top-write read-AUC                   & 0.61 & 0.60 & 0.60 & 0.55 \\
cross-concept write Jaccard          & 0.07 & 0.02 & 0.02 & 0.02 \\
read / write layer                   & 34 / 33 & 24 / 32 & 24 / 32 & 48 / 64 \\
\bottomrule
\end{tabular}
\end{table}

\begin{table}[H]
\centering
\caption{Concept induction via the single-word write target through the attention pathway (12 concepts $\times$ 4 held-out prompts, logit-clean decode, LLM judge vs.\ unsteered baseline; a run counts as success only if the continuation both refers to the concept \emph{and} is at least as fluent as the baseline). Each concept is seeded from one typed word, fracture-expanded into its member set; the write is applied over the band until the concept becomes present, read as raw sign-agreement at the first full-attention band layer's value states (the $v$-sensor), then released once agreement crosses the per-model release threshold (``v9-release''). Per-model settings are the write band (layer span), basin depth $\beta$, coalition size $K$, $v$-sensor layer, and release threshold; the release threshold is tuned per model (a single borrowed value does not travel: Gemma's 0.70 over-writes the others). All twelve concepts share one setting per model.}
\label{tab:control}
\begin{tabular}{lcccc}
\toprule
 & Gemma 3-4B & Mistral 7B & Qwen 3.5-4B & Qwen3-32B \\
\midrule
$D$ & 2560 & 4096 & 2560 & 5120 \\
layers & 34 & 32 & 32 & 64 \\
write band & 8--21 & 14--28 & 12--25 & 24--51 \\
basin depth $\beta$ & 0.3 & 0.6 & 0.4 & 0.5 \\
coalition $K$ & 200 & 200 & 150 & 300 \\
presence sensor (value space) & $v_{9}$ & $v_{15}$ & $v_{16}$ & $v_{25}$ \\
release threshold & 0.70 & 0.50 & 0.60 & 0.85 \\
loop & v9-release & v9-release & v9-release & v9-release \\
\midrule
induced + fluent & \textbf{77\%} & \textbf{77\%} & \textbf{60\%} & \textbf{71\%} \\
\bottomrule
\multicolumn{5}{l}{\footnotesize Shared across every model: spread cutoff $0.5$, floor of 6 writes before release,} \\
\multicolumn{5}{l}{\footnotesize 30 generated tokens. Base-model figures are stable to $\pm2$pp across re-runs;} \\
\multicolumn{5}{l}{\footnotesize the two reasoning-tuned models vary by $\pm10$pp (\S\ref{sec:discussion}).} \\
\end{tabular}
\end{table}

\begin{table}[H]
\centering
\caption{Feature discovery in K/V projections (full-vocabulary negatives, same 50 anchors). Qwen3-4B is substituted for Qwen 3.5-4B because the latter uses hybrid attention, which does not produce standard KV caches at all layers.}
\label{tab:kv}
\begin{tabular}{llccc}
\toprule
Model & Space & Mean per-dim score & Discoverable ($\geq$0.75) & Exceed $p_{95}$ \\
\midrule
\multirow{3}{*}{Gemma 3-4B ($h_{26}$)} & K & 0.887 & 175/175 (100\%) & 175/175 \\
 & V & 0.883 & 174/175 (99\%) & 175/175 \\
 & gate & \textbf{0.924} & 175/175 (100\%) & 175/175 \\
\midrule
\multirow{3}{*}{Qwen3-4B ($h_{24}$)} & K & 0.876 & 175/175 (100\%) & 175/175 \\
 & V & 0.895 & 175/175 (100\%) & 175/175 \\
 & gate & \textbf{0.925} & 175/175 (100\%) & 175/175 \\
\midrule
\multirow{3}{*}{Mistral 7B ($h_{24}$)} & K & \textbf{0.873} & 171/171 (100\%) & 171/171 \\
 & V & 0.842 & 169/171 (99\%) & 171/171 \\
 & gate & 0.851 & 166/171 (97\%) & 171/171 \\
\midrule
\multirow{3}{*}{Qwen3-32B ($h_{48}$)} & K & 0.846 & 172/175 (98\%) & 175/175 \\
 & V & 0.864 & 175/175 (100\%) & 175/175 \\
 & gate & \textbf{0.883} & 175/175 (100\%) & 175/175 \\
\bottomrule
\multicolumn{5}{l}{\footnotesize Qwen3.5-4B is absent because it uses hybrid attention, so per-layer K/V is not well-defined;} \\
\multicolumn{5}{l}{\footnotesize Qwen3-4B carries the claim at the same scale. ``gate'' is the MLP gate activation, \texttt{gate\_act}.} \\
\end{tabular}
\end{table}

\section{Generality: Vision and Audio}
\label{app:generality}

\begin{table}[H]
\centering
\small
\caption{Cross-modality summary. Detect: superclasses (or ESC-50 concepts)
with max per-dim score $>0.70$. Feature AUC: binary superclass feature. Sign-1NN:
Hamming retrieval on signs alone. MLP$-$LogReg: cross-dimension gap. Directed /
shuffled: capture@1 of the write target vs.\ a sign-shuffled target. Full tables
in Appendix~\ref{app:generality}.}
\label{tab:cross_modality_summary}
\begin{tabular}{lccccc}
\toprule
Model (objective) & Detect ${>}0.70$ & Feature AUC & Sign-1NN & MLP$-$LogReg & Directed / shuffled \\
\midrule
DINOv2 (self-sup.)   & 9/12  & 0.831 & 93.0\% & $+$0.007 & --- (no head) \\
ViT-Base (sup.)      & 11/12 & 0.977 & 96.0\% & $-$0.000 & 100\% / 0\% \\
AST (audio, sup.)    & 50/50 & ---   & 97.0\% & $+$0.000 & 100\% / 0\% \\
\bottomrule
\end{tabular}
\end{table}

The experiments above establish per-dim sign structure in language models. We now test whether the same framework applies to transformers trained on entirely different modalities and, crucially, whether it requires classification supervision. We compare a self-supervised and a supervised vision model on the same data, then extend to audio.

\textbf{Models and data.} We evaluate three non-language transformers:
\begin{itemize}
    \item \textbf{DINOv2-Base} \citep{oquab2024dinov2} ($D=768$, 12 layers): self-supervised via self-distillation on LVD-142M with \emph{no classification labels}.
    \item \textbf{ViT-Base} \citep{dosovitskiy2021image} ($D=768$, 12 layers): supervised on ImageNet-1K classification (1000 concepts).
    \item \textbf{AST} \citep{gong2021ast} ($D=768$, 12 layers): supervised on AudioSet classification, evaluated on all 2000 ESC-50 clips (40 per concept) \citep{piczak2015esc}.
\end{itemize}
DINOv2 and ViT-Base share the same ViT-Base architecture ($D=768$, 12 layers) and are evaluated on the same 1000 ImageNet validation images. The only difference is the training objective: self-distillation vs.\ cross-entropy classification. This isolates the effect of supervision on per-dim structure.

\textbf{Method.} We apply the identical per-dim score procedure from \S\ref{sec:method}: extract CLS token hidden states at the final layer ($h_{12}$) for vision models, and mean-pool over all sequence positions for AST (which uses a distillation token rather than a standard CLS for feature extraction). We compute sign patterns and test whether individual dimensions separate semantic concepts.

\textbf{Vision: self-supervised vs.\ supervised on the same data.}

\begin{table}[H]
\centering
\caption{Per-dim sign detection of ImageNet superclasses: DINOv2 (self-supervised, no labels) vs.\ ViT-Base (supervised, 1000-class classification). Same architecture, same evaluation data, different training objective.}
\label{tab:vision_comparison}
\begin{tabular}{lcccccc}
\toprule
 & \multicolumn{3}{c}{DINOv2 (self-supervised)} & \multicolumn{3}{c}{ViT-Base (supervised)} \\
\cmidrule(lr){2-4} \cmidrule(lr){5-7}
Concept & $N$ & Max AUC & Dims${>}$0.70 & $N$ & Max AUC & Dims${>}$0.70 \\
\midrule
Primate & 18 & 0.819 & 30 & 18 & 0.808 & 107 \\
Cat & 13 & 0.780 & 26 & 13 & 0.807 & 48 \\
Insect & 20 & 0.760 & 6 & 20 & 0.779 & 36 \\
Fish & 16 & 0.741 & 10 & 16 & 0.765 & 39 \\
Flower & 11 & 0.736 & 9 & 11 & 0.759 & 29 \\
Musical instr. & 20 & 0.721 & 5 & 20 & 0.802 & 14 \\
Reptile & 34 & 0.719 & 1 & 34 & 0.748 & 12 \\
Bird & 38 & 0.701 & 1 & 38 & 0.744 & 8 \\
Furniture & 12 & 0.749 & 2 & 12 & 0.729 & 6 \\
Vehicle & 33 & 0.667 & 0 & 33 & 0.737 & 3 \\
Dog & 118 & 0.666 & 0 & 118 & 0.758 & 8 \\
Food & 46 & 0.697 & 0 & 46 & 0.679 & 0 \\
\midrule
\multicolumn{1}{l}{Above 0.70} & \multicolumn{3}{c}{9/12} & \multicolumn{3}{c}{11/12} \\
\multicolumn{1}{l}{Feature AUC ($K{=}50$)} & \multicolumn{3}{c}{0.831} & \multicolumn{3}{c}{0.977} \\
\bottomrule
\end{tabular}
\end{table}

Both models exhibit per-dim sign structure. The supervised model achieves higher per-dim score and engages more dims per concept (median 12 vs.\ 5 dims above 0.70), consistent with explicit classification pressure sharpening per-dim specificity. But the structure is already present without supervision: DINOv2 detects 9/12 superclasses from per-dim signs alone, with a 50-dim feature achieving AUC 0.831 for binary animal detection. Classification amplifies per-dim specificity; it does not create the underlying organization.

\textbf{AST: per-dim sign detects audio concepts.}

\begin{table}[H]
\centering
\caption{Per-dim sign detection of ESC-50 audio concepts (AST, $h_{12}$, mean-pool over sequence positions, all 2000 clips, 40 per concept). Showing top-10 of 50 concepts.}
\label{tab:ast_detection}
\begin{tabular}{lcc}
\toprule
Concept & Max dim AUC & Dims $>$ 0.70 \\
\midrule
Train & 0.970 & 69 \\
Thunderstorm & 0.961 & 79 \\
Sea waves & 0.942 & 73 \\
Pouring water & 0.939 & 78 \\
Rooster & 0.939 & 117 \\
Dog & 0.938 & 44 \\
Rain & 0.933 & 58 \\
Sheep & 0.931 & 71 \\
Cow & 0.923 & 68 \\
Chainsaw & 0.921 & 58 \\
\midrule
\multicolumn{3}{l}{\emph{All 50 concepts: 50/50 exceed 0.70; 47/50 exceed 0.80}} \\
\bottomrule
\end{tabular}
\end{table}

All 50 ESC-50 concepts exceed a per-dim score of 0.70, with 47/50 exceeding 0.80, the highest per-dim score across all models and modalities tested. The combination of domain-specific training (AudioSet) and a compact concept space produces near-perfect per-dim discrimination.

\textbf{Sign-only nearest-neighbor retrieval.}

\begin{table}[H]
\centering
\caption{Sign-only 1-NN accuracy (Hamming distance on $\pm 1$ signs, no magnitudes, no training).}
\label{tab:sign_nn}
\begin{tabular}{llccc}
\toprule
Model & Task & Accuracy & Chance & Lift \\
\midrule
DINOv2-Base & Animal vs.\ non-animal & 93.0\% & 60.2\% & +32.8\% \\
ViT-Base & Animal vs.\ non-animal & 96.0\% & 60.2\% & +35.8\% \\
AST & Same superclass (5-class) & 97.0\% & 20.0\% & +77.0\% \\
\bottomrule
\end{tabular}
\end{table}

Pure Hamming distance on sign patterns achieves 93--97\% retrieval accuracy across all models and modalities. The 3\% gap between DINOv2 and ViT-Base reflects the classification sharpening effect, but both far exceed chance with zero learned parameters.

\textbf{Probe ablation: cross-dimension structure is negligible across modalities.}

\begin{table}[H]
\centering
\caption{Probe ablation: LogReg (per-dim, axis-aligned) vs.\ MLP (128 hidden, cross-dim capacity).}
\label{tab:probe_cross_modality}
\begin{tabular}{llccc}
\toprule
Model & Training & LogReg AUC & MLP AUC & Gap \\
\midrule
DINOv2-Base & Self-supervised & 0.931 & 0.938 & $+$0.007 \\
ViT-Base & Supervised & 0.978 & 0.978 & $-$0.000 \\
AST & Supervised & 1.000 & 1.000 & $+$0.000 \\
\bottomrule
\end{tabular}
\end{table}

An MLP with full cross-dimension capacity adds at most $+$0.007 AUC over axis-aligned logistic regression, confirming negligible cross-dim structure regardless of training objective. On the supervised models (ViT-Base and AST), the MLP provides exactly zero benefit. This replicates the language model finding (\S\ref{sec:feature_discovery}) across modalities and supervision regimes.

\textbf{Pairwise MI.} Cross-dimension mutual information: 0.001 bits (DINOv2), 0.001 bits (ViT-Base), 0.002 bits (AST), comparable to language models (0.0003--0.004 bits, Table~\ref{tab:mi}). Only 0.1\% of DINOv2 dim pairs exceed 0.01 bits. Per-dim pairwise independence is modality-universal and does not depend on the training objective.

\textbf{K/V projections preserve per-dim structure across modalities.}

\begin{table}[H]
\centering
\caption{Feature discovery in K/V attention projections for non-language models (same concepts, layer 11). Concepts discoverable = max per-dim score $>$ 0.70.}
\label{tab:kv_cross_modality}
\begin{tabular}{llccc}
\toprule
Model & Space & Discoverable & Exceed null $p_{95}$ \\
\midrule
\multirow{3}{*}{DINOv2-Base (self-supervised)} & H (residual) & 12/12 & 12/12 \\
 & K & 12/12 & 11/12 \\
 & V & 12/12 & 11/12 \\
\midrule
\multirow{3}{*}{ViT-Base (supervised)} & H (residual) & 12/12 & 12/12 \\
 & K & 12/12 & 11/12 \\
 & V & 12/12 & 12/12 \\
\midrule
\multirow{3}{*}{AST (audio, 50 concepts)} & H (residual) & 50/50 & 50/50 \\
 & K & 50/50 & 50/50 \\
 & V & 50/50 & 50/50 \\
\bottomrule
\end{tabular}
\end{table}

The same concepts discoverable from the residual stream are also discoverable from K and V projection dimensions on all three non-language models (Table~\ref{tab:kv_cross_modality}). On AST, all 50 concepts exceed null calibration in both K and V. This confirms that the $W_k$/$W_v$ matrices preserve axis-aligned structure universally, extending the language model finding (\S\ref{sec:attention}) to vision and audio.

\textbf{The write catalog extends to vision and audio.} The results above establish the \emph{read} side cross-modally. We now show that the second catalog, the write target and the control it enables (\S\ref{sec:write_catalog}), extends to non-language transformers as well. For a target class $c$ the write target is the sign of its classifier row, $\boldsymbol{\tau}_c = \mathrm{sign}(W_{\mathrm{cls}}[c])$, the vision/audio analogue of $\mathrm{sign}(\sum_t W_{\mathrm{unembed}}[t])$: a per-dimension polarity read from the frozen head, no training. These models are non-generative classifiers, so ``control'' here means directed movement of the \emph{output decision}, not fluent generation; there is no coherence axis and no closed-loop presence signal (there is nothing to keep fluent), which makes the classifier the cleanest test of the write mechanism in isolation.

\emph{Read and write dimensions are disjoint, cross-modally.} We repeat the disjointness measurement of \S\ref{sec:write_catalog} (rank correlation between per-dim read strength $|\mathrm{AUC}-0.5|$ and write-target polarity), reading concept selectivity across the layer stack. The correlation is depth-gated exactly as the language-model account predicts (\S\ref{sec:discussion}): near zero where the feature is formed but not yet consumed, and rising only at the final layer where the classifier head reads the hidden state directly (ViT: $|\rho|=0.02$ at the early-mid stack, climbing to $0.44$ at $h_{12}$). At a representative layer the correlation and the read-AUC of the top write dimensions are both in the language-model range (Table~\ref{tab:cross_write}), and AST, whose head does not consume the raw final hidden state, is disjoint at every layer. Detection and production are carried by different dimensions of the same basis in vision and audio as in language.

\begin{table}[H]
\centering
\caption{The write catalog on non-language transformers, against the language-model ranges from Table~\ref{tab:write_catalog}. \textbf{Disjoint} $\rho$: rank correlation between per-dim read strength and write-target polarity (read at an early-mid layer, before the head consumes the state). \textbf{Top-write AUC}: mean read-AUC of the top write dimensions (0.5 = chance). \textbf{Jaccard}: cross-concept write-target overlap.}
\label{tab:cross_write}
\begin{tabular}{lccc}
\toprule
 & read--write $\rho$ & top-write AUC & cross-concept Jaccard \\
\midrule
Language models (4)  & 0.01--0.11 & 0.55--0.61 & 0.02--0.07 \\
ViT-Base             & 0.02       & 0.54       & 0.05 \\
AST                  & 0.01       & 0.54       & 0.04 \\
\bottomrule
\end{tabular}
\end{table}

\emph{The write target steers the output decision.} We inject $\boldsymbol{\tau}_c$ through the same attention output pathway as in language (\S\ref{sec:write_catalog}): a coalition of \texttt{o\_proj}-input coordinates whose weight-column signs agree with $\boldsymbol{\tau}_c$, pushed across a write band of layers, with basin depth $\beta$. On real inputs that are \emph{not} the target class (so the write must override existing content), the injection drives the classifier to the target: capture rises monotonically with $\beta$ and saturates at $100\%$ on both models (12/12 ImageNet, 12/12 AudioSet concepts), while a sign-shuffled target of identical coalition size and $\beta$ stays at chance at every strength (Table~\ref{tab:cross_control}). The effect is thus carried by the sign \emph{pattern} of $\boldsymbol{\tau}_c$, not by the magnitude of the push. Because the classifier is non-generative we report the unsteered forward pass (no injection) as the baseline; the write is a genuine intervention rather than a decode-time heuristic.

\begin{table}[H]
\centering
\caption{Directed output control via the write target through the attention pathway, on real non-target inputs (ViT: 40 ImageNet images $\times$ 16 classes; AST: 30 ESC-50 clips $\times$ 12 AudioSet classes). \textbf{Directed}: capture@1 with $\boldsymbol{\tau}_c$. \textbf{Control}: sign-shuffled target, matched coalition size and $\beta$, 3 seeds. \textbf{Specificity}: mean on-target vs.\ off-target capture across the concept $\times$ concept matrix. \textbf{Amplifier}: capture using only the coordinates shared across concepts (the generic amplifier) vs.\ with those coordinates removed.}
\label{tab:cross_control}
\begin{tabular}{lccccc}
\toprule
 & Directed & Control & On/off-target & Amplifier-only & Amplifier-removed \\
\midrule
ViT-Base & 100\% & 0\% & 100\% / 0\% & 0\% & 100\% \\
AST      & 100\% & 0\% & 100\% / 0\% & 2\% & 100\% \\
\bottomrule
\end{tabular}
\end{table}

The control is specific and concept-carried, not a generic push. Steering toward one class raises only that class to the top (on/off-target $100\%/0\%$; a rank readout shows genuine near-neighbours, e.g.\ tiger and tiger-cat, raised together while unrelated classes are untouched, so the specificity tracks semantics rather than being merely numerical). And the signal lives in the concept-specific coordinates: removing the coordinates shared across all concepts (the ``generic amplifier'' of \S\ref{sec:write_catalog}) leaves capture at $100\%$, whereas steering with \emph{only} those shared coordinates fails entirely and collapses to the dataset's most frequent labels. Finally, the control is a distributed band, not head-adjacency: no single sub-band suffices (on AST, early, middle, and late thirds each reach at most $8\%$ alone), while the full band reaches $100\%$; the concept consensus is assembled across depth, as in language (\S\ref{sec:discussion}).

\textbf{Summary.} Three training objectives, next-token prediction (language), self-distillation (DINOv2), and classification (ViT-Base, AST), all produce the same per-dim sign organization. Both catalogs port across modalities: residual-stream detection and K/V preservation are equally strong, and the write target, read from the frozen head, is disjoint from the read catalog and steers the output decision through the attention pathway just as it steers generation in language. The shared factor is the architecture: a residual stream fed by additive attention and FFN writes.

\section{Lifecycle: Early-Write and Late-Read Details}
\label{app:circuit}
Per-layer curves for the two lifecycle measurements of
Section~\ref{sec:detect-before-read} on all eight models (Gemma 3-4B, Qwen 3.5-4B,
Mistral 7B, Qwen3-32B, Mamba-2.8B, OLMoE-1B-7B, Pythia-1.4B, RWKV-4-3B), run at
float32. \emph{Early write:} restoring only the clean sign pattern, with the
corrupted magnitudes kept, at a single early layer recovers a median 94--100\% of
the corrupted answer probability within the first 15\% of depth, while restoring
random signs pushes the answer below its corrupted level. \emph{Late read:}
sign-vote answer accuracy against the concept addresses reaches 68--80\% top-1
and peaks at 74--89\% depth. The per-layer curves and controls behind both measurements follow the
same protocol as \S\ref{sec:detect-before-read}. Table~\ref{tab:sign_restore} gives the early-write
recovery per model across depth: it is near-total in the earliest layers and fades as the answer
moves off the subject position.

\begin{table}[H]
\centering
\caption{Early write, per model (float32). Median fraction of the destroyed answer probability
recovered by restoring only the clean sign pattern at the subject position, with the corrupted
magnitudes kept, at the layer nearest each fraction of depth. The last column restores random signs
instead, at the model's best early layer; values below zero mean the answer falls below its
corrupted level.}
\label{tab:sign_restore}
\begin{tabular}{lcccccc|c}
\toprule
 & & \multicolumn{5}{c|}{sign restore, by depth} & random \\
Model & Layers & 5\% & 15\% & 25\% & 50\% & 75\% & signs \\
\midrule
Gemma 3-4B   & 34 & 0.98 & 0.96 & 0.97 & 0.86 & 0.11 & $-$0.42 \\
Qwen 3.5-4B  & 32 & 0.96 & 0.94 & 0.84 & 0.60 & 0.03 & $-$0.01 \\
Mistral 7B   & 32 & 1.00 & 0.93 & 0.93 & 0.54 & 0.07 & $-$2.00 \\
Qwen3-32B    & 64 & 0.96 & 0.83 & 0.93 & 0.83 & 0.47 & $-$0.87 \\
\midrule
Mamba-2.8B   & 64 & 1.00 & 1.00 & 0.99 & 0.97 & 0.01 & $-$0.07 \\
OLMoE-1B-7B  & 16 & 0.97 & 0.96 & 0.95 & 0.93 & 0.18 & $-$0.12 \\
Pythia-1.4B  & 24 & 0.93 & 0.94 & 0.92 & 0.57 & 0.00 & $-$0.22 \\
RWKV-4-3B    & 32 & 0.99 & 0.96 & 0.93 & 0.60 & 0.17 & $-$0.06 \\
\bottomrule
\end{tabular}
\end{table}

\begin{table}[H]
\centering
\caption{Early read, per model (float32). AUC of the harm-type sign pattern at the last prefill
token, at the layer nearest each fraction of depth: harmful vs.\ benign prompts / harmful vs.\
hard-negative prompts (benign requests with charged wording). 10 prompts per group, so each AUC is
coarse (0.01 steps) and single layers fluctuate; the gate reads early on every architecture.}
\label{tab:harm_depth}
\begin{tabular}{lcccccc}
\toprule
Model & 6\% & 12\% & 19\% & 25\% & 50\% & 75\% \\
\midrule
Gemma 3-4B   & 0.92 / 0.87 & 0.98 / 0.92 & 0.95 / 0.93 & 0.95 / 0.89 & 0.93 / 0.89 & 1.00 / 0.99 \\
Qwen 3.5-4B  & 0.83 / 0.65 & 0.95 / 0.82 & 0.92 / 0.86 & 0.97 / 0.91 & 0.92 / 0.90 & 0.98 / 0.93 \\
Mistral 7B   & 0.93 / 0.91 & 0.86 / 0.84 & 0.92 / 0.94 & 1.00 / 0.94 & 1.00 / 0.98 & 1.00 / 0.95 \\
Qwen3-32B    & 0.93 / 0.70 & 0.99 / 0.93 & 0.92 / 0.84 & 0.97 / 0.86 & 0.94 / 0.94 & 0.93 / 0.88 \\
\midrule
Mamba-2.8B   & 0.97 / 0.87 & 0.81 / 0.78 & 0.97 / 0.93 & 0.92 / 0.89 & 1.00 / 0.98 & 0.99 / 0.97 \\
OLMoE-1B-7B  & 0.94 / 0.89 & 0.99 / 0.95 & 1.00 / 0.91 & 1.00 / 0.96 & 0.97 / 0.94 & 1.00 / 0.95 \\
Pythia-1.4B  & 0.99 / 0.82 & 0.94 / 0.88 & 0.98 / 0.90 & 0.99 / 0.91 & 0.97 / 0.95 & 0.96 / 0.95 \\
RWKV-4-3B    & 0.92 / 0.78 & 0.94 / 0.79 & 0.86 / 0.76 & 0.96 / 0.92 & 0.98 / 0.95 & 0.99 / 0.96 \\
\bottomrule
\end{tabular}
\end{table}

\begin{figure}[h]
\centering
\includegraphics[width=0.62\linewidth]{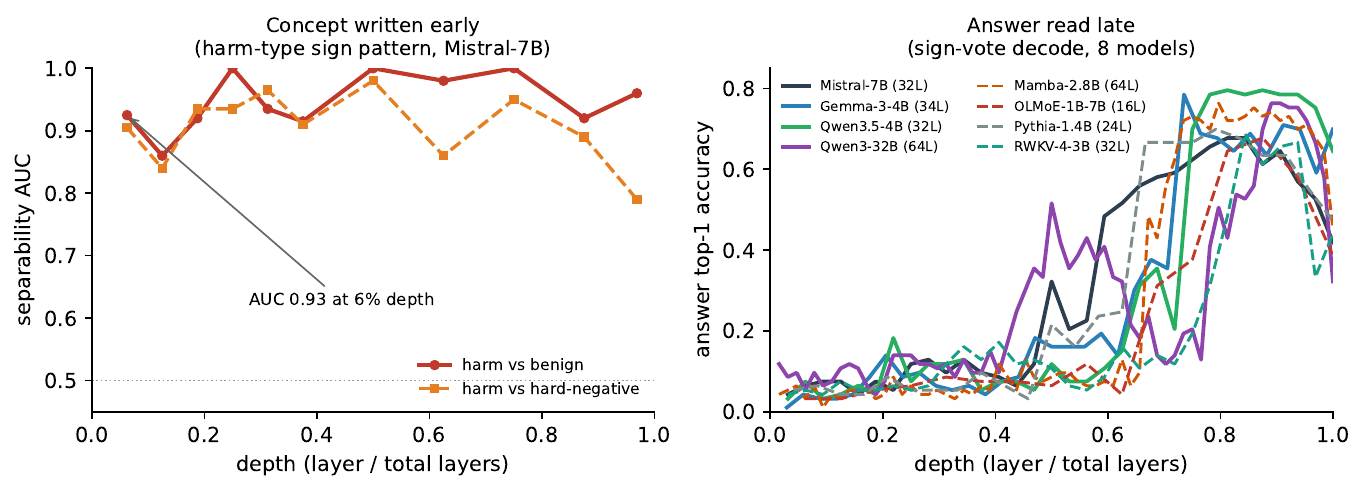}
\caption{The two ends of the lifecycle on a depth axis. \emph{Left:} early
detectability, the harm-type sign pattern separating harmful from benign prompts
at AUC 0.93 by 6\% depth (Mistral-7B; all eight models in Table~\ref{tab:harm_depth}), the signal
the prefill guardrail of \S\ref{sec:guardrail} reads. \emph{Right:} late readability, sign-vote answer
accuracy against the concept addresses, which peaks only at 74--89\% of depth
across eight models (solid: the four language models; dashed: Mamba, OLMoE,
Pythia, RWKV). The causal early-write result (sign-restore recovery)
is given per model in Table~\ref{tab:sign_restore} rather than plotted here.}
\label{fig:early_late}
\end{figure}

\section{Sign-Lens Neighbor Sets}
\label{app:evidence}
Full nearest-neighbor sets and per-depth listings behind the sign-lens ladder
of Table~\ref{tab:fracture_ladder_full}, together with the guardrail and monitor
operating details (threshold sweeps, per-model curves, and the audited
top-weighted dimensions), are obtained with the same Sign Lens reads described
in \S\ref{sec:signlens}.

\begin{table}[h]
\centering
\footnotesize
\setlength{\tabcolsep}{4pt}
\begin{tabular}{@{}l l l l l@{}}
\hline
Prompt (answer) & 12\% depth & 38\% depth & 50\% depth & 75\% depth \\
\hline
\emph{capital of France} (Paris)
 & capital, city, currency
 & London, Madrid, Paris
 & Spain, France, Italy
 & \textbf{Paris}, France, Europe \\
\emph{capital of Japan} (Tokyo)
 & capital, city, cities
 & Delhi, London, Tokyo
 & Japan, Spain, Germany
 & \textbf{Tokyo}, Japan, Cuba \\
\hline
\end{tabular}
\caption{The Sign Lens on two factual prompts (Mistral-7B, nearest neighbors of the
last-position sign pattern by sign agreement; latin-script tokens shown). Both show
the same ordering: the capital-and-city type is coherent by 12\% depth, candidate
capital names appear mid-stack, and the answer token dominates only near 75\%.}
\label{tab:fracture_ladder_full}
\end{table}

\end{document}